\documentclass{article}

\usepackage{arxiv}

\usepackage{amsmath,amssymb,mathrsfs}

\usepackage{array}

\usepackage{stfloats}

\usepackage{url}

\usepackage{comment}
\usepackage[inline]{enumitem}
\usepackage{lineno,hyperref}
\usepackage{siunitx}
\usepackage{booktabs}
\usepackage{multirow}

\usepackage[utf8]{inputenc} 
\usepackage[T1]{fontenc}    

\usepackage{amsfonts}       
\usepackage{nicefrac}       
\usepackage{microtype}      

\usepackage{graphicx}
\usepackage{natbib}
\usepackage{doi}

\title{Large Scale Passenger Detection with Smartphone/Bus Implicit Interaction and Multisensory Unsupervised Cause-effect Learning}

\author{Valentino~Servizi\\
	Department of Technology, \\
	Management and Economics\\
    Technical University of Denmark (DTU)\\
	\texttt{valse@dtu.dk} \\
	\AND
	Dan R.~Persson \\
	Department of Applied Mathematics \\
	and Computer Science\\
    DTU\\
	\And
	Francisco C.~Pereira \\
	Department of Technology, \\
	Management and Economics\\
    DTU\\
	\And
	Hannah~Villadsen \\
	Department of People and Technology\\ 
	Roskilde University\\
    Denmark\\
	\And
	Per~Bækgaard \\
	Department of Applied Mathematics \\
	and Computer Science\\
    DTU\\
    \And
	Jeppe~Rich\\
	Department of Technology, \\
	Management and Economics\\
    DTU\\
	\And
	Otto A.~Nielsen\\
	Department of Technology, \\
	Management and Economics\\
    DTU\\
}

\renewcommand{\shorttitle}{Cause-effect Learning}

\hypersetup{
pdftitle={BIBO Cause-effect Learning},
pdfsubject={Arxive},
pdfauthor={Valentino~Servizi et al.},
pdfkeywords={Device-to-device, Sensor-to-sensor, Ground-truth-validation, Wasserstein-auto-encoders, Autonomous-vehicles},
}

\begin{document}
\maketitle

\begin{abstract}
Intelligent Transportation Systems (ITS) underpin the concept of Mobility as a Service (MaaS), which requires universal and seamless users' access across multiple public and private transportation systems while allowing operators' proportional revenue sharing.
Current user sensing technologies such as Walk-in/Walk-out (WIWO) and Check-in/Check-out (CICO) have limited scalability for large-scale deployments. These limitations prevent ITS from supporting analysis, optimization, calculation of revenue sharing, and control of MaaS comfort, safety, and efficiency. We focus on the concept of implicit Be-in/Be-out (BIBO) smartphone-sensing and classification. 

To close the gap and enhance smartphones towards MaaS, we developed a proprietary smartphone-sensing platform collecting contemporary Bluetooth Low Energy (BLE) signals from BLE devices installed on buses and Global Positioning System (GPS) locations of both buses and smartphones. To enable the training of a model based on GPS features against the BLE pseudo-label, we propose the Cause-Effect Multitask Wasserstein Autoencoder (CEMWA). CEMWA combines and extends several frameworks around Wasserstein autoencoders and neural networks. As a dimensionality reduction tool, CEMWA obtains an auto-validated representation of a latent space describing users' smartphones within the transport system. This representation allows BIBO clustering via DBSCAN.

We perform an ablation study of CEMWA's alternative architectures and benchmark against the best available supervised methods. We analyze performance's sensitivity to label quality. Under the naïve assumption of accurate ground truth, XGBoost outperforms CEMWA. Although XGBoost and Random Forest prove to be tolerant to label noise, CEMWA is agnostic to label noise by design and provides the best performance with an 88\% F1 score.

\end{abstract}

\keywords{Device-to-device \and Sensor-to-sensor \and Ground-truth-validation \and Wasserstein-auto-encoders \and Autonomous-vehicles}

\section{Introduction}
%
%
%
%
Tracking passenger movements through the public transport network, seamlessly and without direct human interaction, requires accurate models and methods to discriminate between passengers that are using the public transport network and anyone else outside the transport network.
While the accurate solution of such an implicit Be-In/Be-Out (BIBO) classification problem \citep{narzt2015a}, is directly relevant as a mean to collect important data from the public transport system, e.g. Check-in/Check-out or Walk-in/Walk-out statistics, it is relevant for other areas as well. This includes as an example, the tracking of persons entering buildings to comply with safety measures and the registration, and tracking of people in supermarkets to support crew management in different parts of the supermarket. However, tracking of public transport users represent a more complex problem in that buses and passengers move in space- and time. As a result, we will argue that the ability to provide robust solutions for public transport applications is a stepping-stone for these other relevant applications.

Solving the before mentioned classification problem is important for several reasons. Firstly, on the very practical side it provides a means to collect valuable data about passenger flows that would otherwise have been lost for users paying by cash, or accidentally traveling without checking-in. Secondly, it would enable context-aware surveying and services while lifting the burden of explicit interaction from passengers. Thirdly, for planning optimal departure times and routes of a trip through the public network, it would support personalized dynamic recommendations.

In a wider perspective the presented methodology can be seen as an important component in Mobility-as-a-service (MaaS) systems. MaaS combines multiple transport modes as transport services--e.g., car, bus, bike, scooter--offered through a single interface, and paid with the same unique subscription, as the media contents on ``Netflix'' \citep{hietanen2014mobility,HENSHER2021172}. Hence, MaaS is essentially \textit{``a data-driven, user-centered paradigm, powered by the growth of smartphones''} \citep{goodall2017rise}. Regardless from the perspective, MaaS ultimate goal is to enable a door-to-door public service, attractive for the passengers, and competitive with, e.g., privately owned cars. 
In this context, the ability to accurately track passengers while traveling would underpin the efficient capacity planning for a dynamic, responsive, and intelligent public transport paradigm.

In the MaaS context,  smartphone-based automatic fare collection systems (AFCS) with BIBO could allow the integration of public service ticketing, automatic price calculation, and a fair cost split across multiple operators. The latter point includes emerging providers of, e.g., car- and bike-sharing services. Compared to CICO and WIWO, BIBO offers at least two advantages:
\begin{enumerate*}[label=(\roman*)]
    \item public transit increased comfort for passengers \citep{wirtz2019a}; and
    \item operational integration mostly software, with a negligible impact on new physical infrastructure.
\end{enumerate*}
The second point means potentially lower access barriers for emerging transport service providers to MaaS. For the first, we refer to the passengers increased comfort with the term ticketless. Ticketless identifies the perspective of a system ability to flexibly adapting the transport service bill to the user's journey(s) across multiple service providers, as opposed to the perspective of multiple tickets necessary from multiple service providers, for the same journey.

From the Big Data perspective, handling this binary classification problem with supervised machine learning methods presents the following challenges:
\begin{enumerate}
    \item Controlling noise in the labels:
    \item Operating a sustainable labels collection cost;
    \item Minimizing the impact of sensors and data collection on the battery; and
    \item Minimizing the users' privacy exposure.
\end{enumerate}
These challenges involve the service operator's perspective in the first case and the smartphone user's perspective in the others.

Although from a ticketing perspective there should be no noise, thus one should only be charged when he or she uses a transport service, when using tickets as labels to train machine learning algorithms, the assumption of possible undetected ticketing errors from both sides--passenger and service provider--seems more than reasonable.


Mining transport behavior from smartphones data relies, among other sensors, on Global Positioning System (GPS), Inertial Navigation System (INS), and Bluetooth Low Energy (BLE) signal\citep{servizi2020a}. In urban areas, where 80\% of public transport demand occurs \citep{TUreport2020} (e.g., in Denmark), the classification of sensors' observations is complex. With GPS, any transportation mode looks the same due to a combination of factors, such as GPS errors in urban canyons, proximity between pedestrians and buses, and vehicles' low speeds in congested traffic \citep{cui2003a}. With INS, multiple habits, each corresponding to whether one carries a smartphone, e.g., in the pocket or the bag, determine different sensors patterns \citep{Wang2019}; the integral of any noise included in the sensors' signal, in addition, leads to often unmanageable error drifts \citep{Foxlin1996}. The BLE signal, which is extensively studied for indoor tracking, presents an excellent potential for proximity sensing and battery efficiency \citep{bjerreNielsen2020a}. However, smartphones' signal records of BLE devices in proximity suffer from signal gaps \citep{Malmberg_2014}; a higher spatial density of BLE devices allows good indoor-tracking performance, but such a density is not scalable at a city scale. In contrast, GPS and INS scaling potential correspond to a heavy impact on the smartphones' battery \citep{servizi2020a}. In the first case, the sensor is directly responsible for the energetic consumption. In the second case, the sensors' energy consumption is sustainable as long as the signals are classified online within the smartphone. Yet, due to the high sampling rate necessary for achieving acceptable classification performance, $>\SI{20}{\hertz}$, data consumption outside the smartphone would imply high network energy consumption for data transfer \citep{servizi2020a}. 
In the assumption of training a supervised machine learning algorithm with high-quality labels, BIBO binary classification in the urban context seems a difficult task. When labels' quality degrades, we face another limitation as classifiers' performance can be highly biased—consequently, decisions would be based on scores looking high when they are low in reality and vice-versa \citep{servizi2021context}.
To overcome the limitations mentioned above, in this work, we rely on a unique dataset collected during three months of autonomous buses' operations across a local public network in Denmark. The dataset includes the GPS and BLE trajectories collected from buses and passengers' smartphones through a proprietary smartphone-sensing platform, including 300 BLE devices installed in buildings near the bus network, in the buses, and at bus stops. Another set of the data provides high-quality ground truth collected by users that followed precise instructions on individual sequences of origins and destinations within the bus network, along specific routes \citep{shankari2020mobilitynet}.

\subsection{Literature Review}
The solution we propose for the BIBO classification problem involves the implicit interaction of passenger smartphones, buses, and bus-network \citep{servizi2021context, narzt2015a}. Therefore, it falls within the intersection of several disciplines converging around smartphone-based travel surveys and smartphones indoor tracking with BLE network interaction. In the first case, leveraging smartphone onboard sensors, we are interested in the limitations of the methods for mode detection in general and bus detection in particular \citep{wirtz2019a}; in the second case, we are interested in how to deal with BLE signals \citep{servizi2021context}.

The literature on mode detection from smartphones data is pervasive. GPS and INS sensors are the most used also to provide location- and person-agnostic mode classification. GPS and INS systems generate very different trajectories. The first system provides a geospatial time series with a sampling rate $\ge \SI{1} {\hertz}$  \citep{StopDetection, Dabiri2018}; the second system, a three-dimension time series along the three axes of the smartphone's reference frame, and a sampling rate $\ge \SI{20}{\hertz}$ \citep{servizi2020a, cornacchia2017a}. To prepare the data for the classification, the steps one follows to clean and segment these trajectories differ too. However, the best-performing classification methods consist of two main groups. The first group includes supervised methods, such as decision trees, random forest, and XGBoost \citep{Koushik2020}; the second group has various configurations of artificial neural networks (ANN), both supervised and semi-supervised. Unsupervised methods based on clustering are applied directly to features extracted from GPS and INS, but their performance seems below the supervised and semi-supervised methods mentioned above. The blooming literature on both GPS- and INS-based mode detection proposes very effective methodologies, equally accurate when datasets include urban and outskirt areas and multiple transportation targets \citep{servizi2020a}. However, at low speeds, state-of-the-art INS-based online classifiers available on the leading smartphone operation systems seem unable to discriminate between bus and walk mode. In contrast, GPS and BLE classifiers show higher performance \citep{servizi2021context}.

Among the studies focusing on mode detection and public transportation, specifically buses, the most promising are considering the interaction between users and the transport network.  This interaction could be expressed as the time series of the distances between each point of a smartphone's GPS trajectory and each point of interest (PoI) extracted from the infrastructure mapped on GIS \citep{Semanjski2017}. The classification could be point-based, thus relying on short segments. Another approach, which we define segment-based \citep{servizi2020a}, could look at longer trip segments and the periodicity of stops typical of any bus operation \citep{zhang2011multi}. However, while the first approach suffers the limitation from the GPS error in dense urban areas, the second approach seems ineffective for short trips.

Literature focusing on BLE and WiFi signals--both based on the same communicaiton frequency and protocols sharing some similarities--converges between indoor tracking and mode detection. The traditional methodologies leverage the Friis equation, and the trilateration \citep{kotanen2003a, subhan2011a}. However, machine learning methods such as random forests and Gaussian processes are effective in BLE or WiFi fingerprint classification, and spatial signal mapping \citep{chen2015a, subhan2013a, p2012a}. To allow optimal BIBO sensing and classification with BLE devices, we find no clear contributions on the minimum spatial density of BLE devices, nor how to cover the scale of a city \citep{servizi2021context}. Therefore, we rely on literature about indoor tracking \citep{yassin2017a} and preliminary BIBO experiments with BLE signals \citep{servizi2021context}, suggesting that BLE devices installed in buses and bus stops could offer a coverage sufficient for classification. Consequently, such a configuration would have the potential to cover the entire city at a reasonable cost.

The parallel growth of computation power and data volume kept in check the tradeoff between computational capacity and classification performance. On the one hand, Computation Processing Units (CPU) and Graphical Processing Units (GPU) have created sizeable extra computation potential. On the other hand, the pursuit of better accuracy leveraging, for example, the pervasive introduction of cheap sensors and rich Geographic Information Systems (GIS), immediately absorbed this additional capacity. Overall, transportation mode classifiers deployed on data from urban and densely populated areas did not increase their performance proportionally with the data consumption. Therefore, statistical methods developed before the Big Data paradigm \citep{sss2009}, and machine learning methods developed after \citep{Koushik2020}, may still compete. A factor emerging from the literature is that methods still depend heavily on labels. Even though some semi-supervised configuration of artificial neural networks exists in this field and reduces the need for labels in the classifier's training phase, filtering a subset of high-quality labels from Big dataset is still very challenging and hardly scalable. For example, continuous disruptions of transport operations due to roadwork or special events would also disrupt any classifier trained with labels that no longer reflect the transport network \citep{petersen2021short}. Even in the assumption of operations stability, the impact of flipping and overlaying labels--potentially present due to human collection errors--seems still critical. Supervised classifiers deployed on time series, e.g., for the BIBO task, could deliver biased classifications and threaten the system's sustainability at scale. The problem deserves more attention in this field, and for time series requires at least the same attention granted to independent and identically distributed data. Systematic studies and appropriate methodologies in the second case exist, such as for image classification. However, for time series classification these contributions are only partially applicable. Furthermore, existing preliminary studies about the impact of flipping labels on time series classification show that severe bias on the measurements of these classifiers' performance is present when just 10\% of the labels are wrong. In such a case, although the classifiers might be resilient to labels' noise, analysts and practitioners would base their decisions on a biased performance evaluation, simply because the error rate in human validated labels is unknown \citep{servizi2021context}.

\subsection{Contribution of the Paper}
This paper focuses on the combined use of GPS and BLE signals for unsupervised autovalidated BIBO classification of bus passengers. Representing the user via the smartphone and the bus via a BLE device, we use sensors signals as pseudo labels to learn discriminating when a user is inside (BI) or outside (BO) the bus.

The central intuition is that when the user is inside the bus (BI) the distance between smartphone and bus should be close to zero, and the proximity to BLE devices installed in the bus would cause the highest signal strength. Vice-versa, when the user is outside the bus (BO), the considerable distance between the user and the BLE device should cause the lowest signal strength or no signal at all.

To learn the cause-effect relationship between smartphone-bus proximity and BLE signal strength, we implement two parallel Wasserstain Autoencoders (WAE). One learns how to reconstruct the time series of the BLE signal (effect) given the smartphone-bus proximity (cause). Given the BLE signal strength (effect), the other learns to rebuild the smartphone-bus distance (cause). 
We define this configuration as a cause-effect multi-task Wasserstein Auto-encoder (CEMWA). From the unsupervised training of this CEMWA, we learn to reduce the description of the interaction between passengers and buses to only four dimensions. In this 4-dimensional latent space, the observations self-organize such that discrimination between BI and BO classes is possible through unsupervised clustering with Density-based spatial clustering of applications with noise (DBSCAN).

CEMWA combines and extends the following frameworks.
\begin{enumerate*}[label=(\roman*)]
    \item Split-brain Auto-encoder configuration by \cite{zhang2016splitbrain};
    \item Deep clustering for unsupervised learning  \citep{Caron_2018_ECCV};
    \item Multi-task formulation of the objective function by \cite{kendall2018multi};
    \item Maximum Mean Discrepancy (MMD) formulation of the objective function for generative models by \cite{gretton2008kernel}; and 
    \item MMD extension to Wasserstein Auto-encoders by \cite{tolstikhin2017wasserstein}.
\end{enumerate*}

The resulting architecture solves the scalability problem related to noise in labels. We perform an ablation study including traditional WAE architectures and supervised methods. Results show that our unsupervised classifier solves the negative impact of the label-induced bias affecting supervised classifiers. Moreover, the architecture we propose embodies a solution for signal data imputation, which is generally a critical and separate step necessary to perform good classification. Finally, since the method relies only on the interaction between smartphone and bus, temporary or permanent disruptions of the network would not affect the classification task.

\begin{figure}
\centering
\begin{minipage}{0.8\textwidth}
  \centering
  \includegraphics[width=\textwidth]{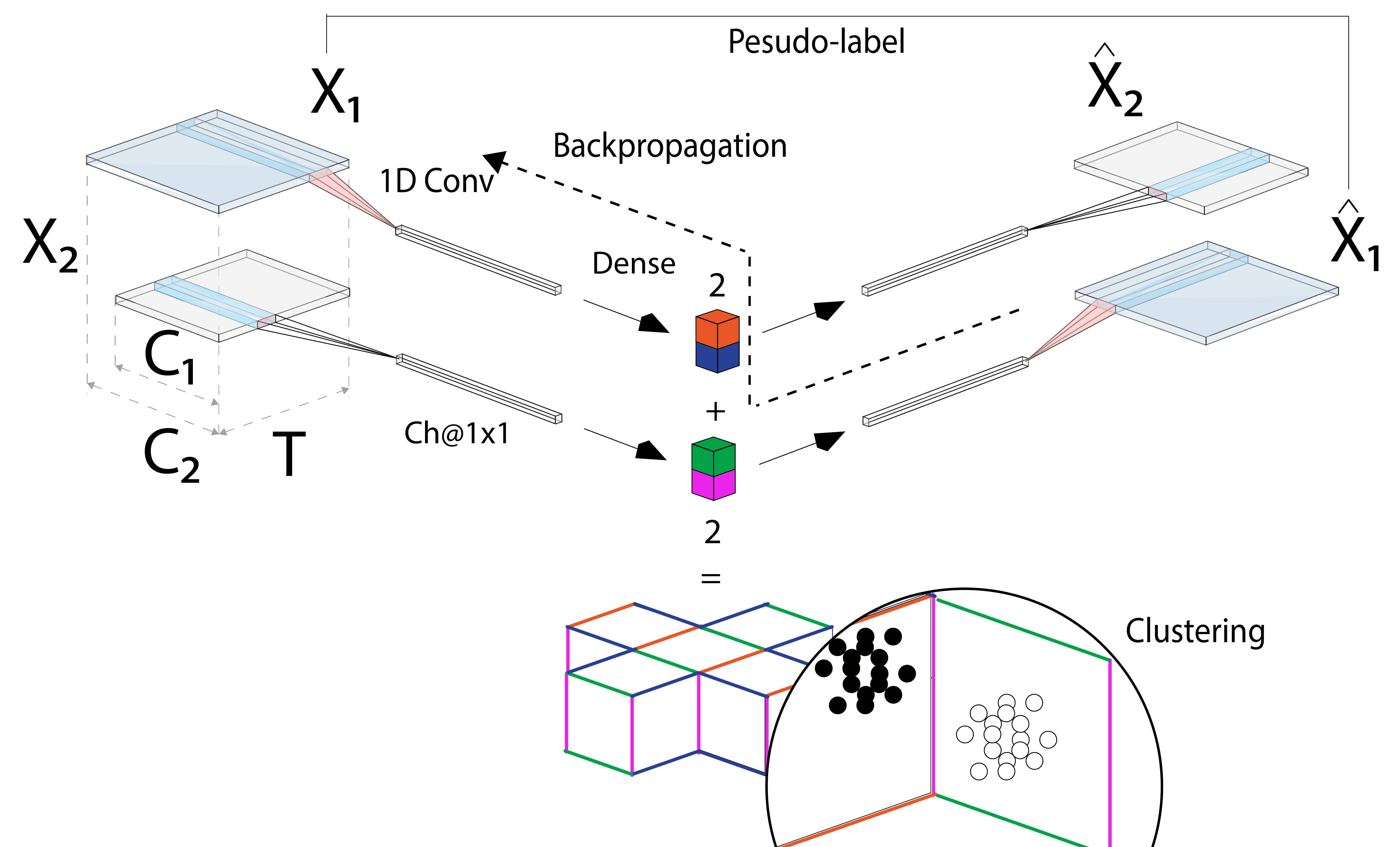}
  \caption{Cause-effect Multi-task Wasserstein Auto-encoder (CEMWA) independent cross-reconstruction of $X_1, X_2$  minimizing \eqref{eq:CEMWA} and clustering of the resulting latent space, 5028 parameters.
  \newline
  }
  
  \label{fig:CEMWA}
\end{minipage}%

\begin{minipage}{.8\textwidth}
  \centering
  \includegraphics[width=\textwidth]{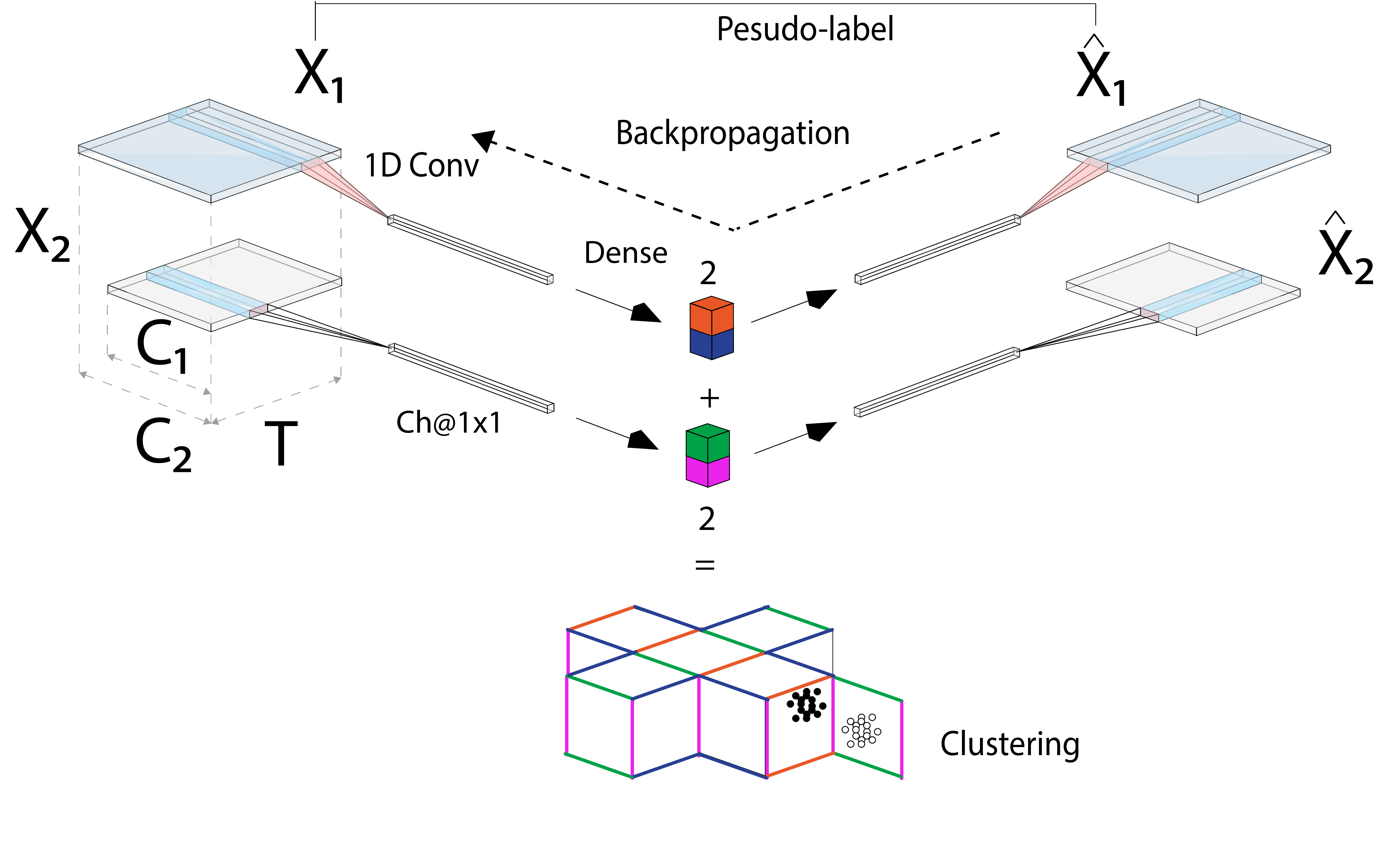}%
  \caption{Multi-task Wasserstein Auto-encoder (MWA) independent reconstruction of $(X_1, X_2)$  minimizing \eqref{eq:multi-task-kendal}, with $c=\mathcal{L}_{\textit{WAE}}$ and clustering of the resulting latent space, 5028 parameters.
  }
  
  \label{fig:MWA}
\end{minipage}

\end{figure}

\begin{figure}
\centering
\begin{minipage}{.8\textwidth}
  \centering
  \includegraphics[width=\textwidth]{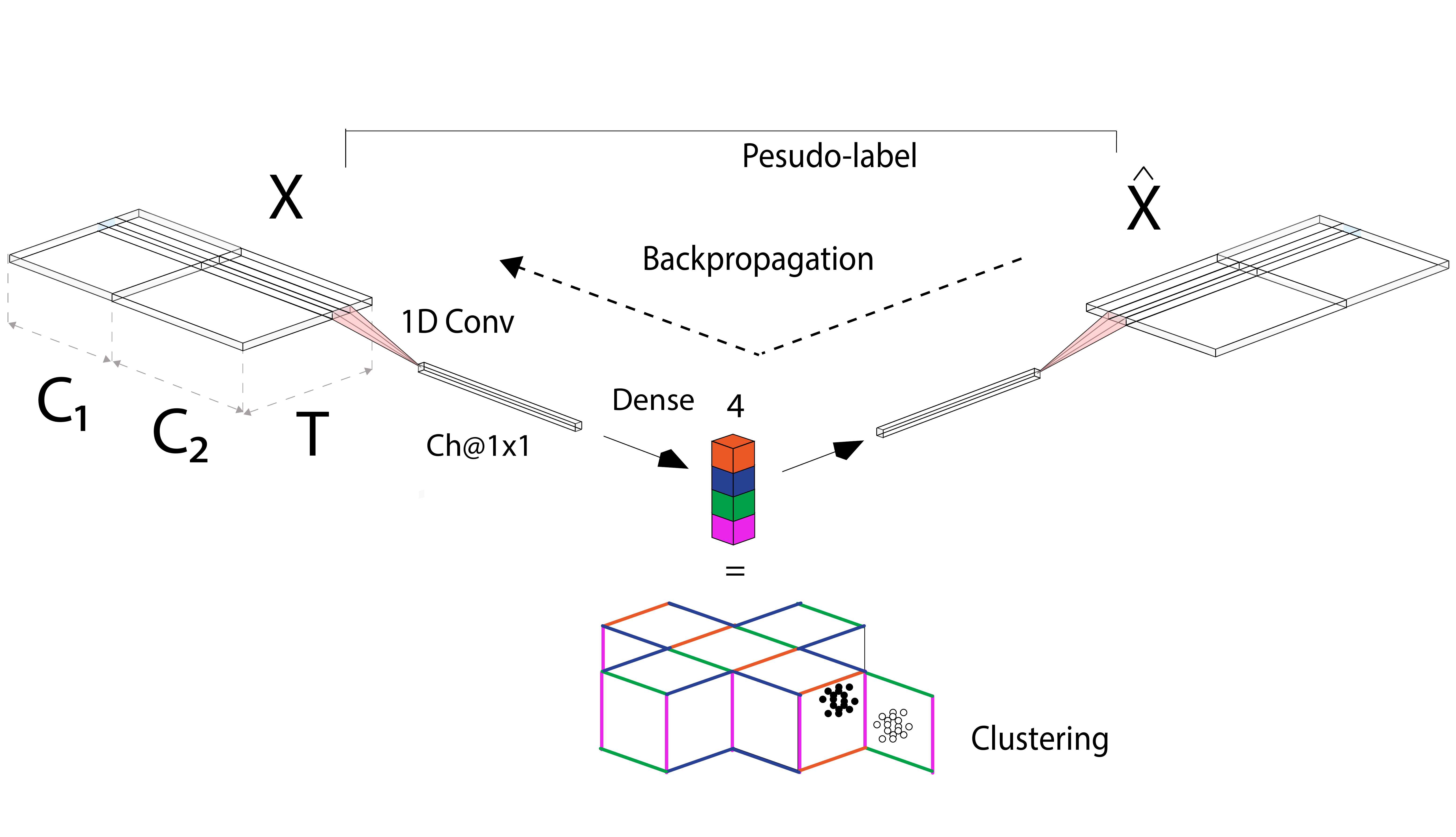}%
  \caption{Wasserstein Auto-encoder (WA) reconstruction of $X = (X_1, X_2)$ minimizing \eqref{eq:tolstikhin2017wasserstein} and clustering of the resulting latent space, 4932 parameters.  
  }
  
  \label{fig:WA}
\end{minipage}
\end{figure}

\section{Methods and Materials}
\label{sec:methodsNmaterials}
This section presents a number of frameworks supporting our goal of substituting ordinary labels for training supervised or semi-supervised artificial neural networks specialized in processing GPS signal. Three are the main steps behind the intuition. Firstly, instead of labels we leverage an independent sensor time-series--BLE--for representation learning of cause-effect relationship between GPS and BLE. Secondly, to avoid confounding correlations between the two sensors' signals, we design and fine-tune a specific encoder-decoder architecture based on a general formulation of regularized auto-encoders. Lastly, with DBSCAN, we turn into classes the representations learned via independent sensors time-series--GPS and BLE.

Following the notation of \cite{tolstikhin2017wasserstein}, we identify sets with calligraphic letters (i.e. $\mathcal{X}$), random variables with capital letters (i.e. $X$)., and values with lower case letters (i.e. $x$).

Let $X \in \mathbb{R}^{t \times d}$ be the tensor describing the smartphone/bus interaction, in a time window of $t$ observations, which $d$ independent feature channels express such that: $X_1 \in \mathbb{R}^{t \times d_1}$ represents the channels deriving from the GPS sensors; $X_2 \in \mathbb{R}^{t \times d_2}$, from the BLE devices network; where $(X_1, X_2) = X$ and $\mathcal{D}_1 \bigcup \mathcal{D}_2 \subseteq \mathcal{D}$, with $|\mathcal{D}| = d$ .

We would like to learn a representation for $X$ solving the prediction problem $\hat{X} = (\hat{X_1}, \hat{X_2}$), where $\hat{X_1} = \mathcal{F}_1(X_2)$, and $\hat{X_2} = \mathcal{F}_2(X_1)$. $\mathcal{F}_1$ learns the cause-effect relationship between smartphone-bus proximity and BLE signal strength, while $\mathcal{F}_2$ learns the inverse cause-effect relationship of the same interaction between smartphone and bus.

$\mathcal{F}$ represents a class of non-random generative Encoder/Decoder models determinalistically mapping input points to the latent space with a convolutional neural network (CNN) via Encoder, and latent codes to output points with a transpose CNN via Decoder. To learn $\mathcal{F}$, we minimize the Wasserstein optimal transport cost \eqref{eq:tolstikhin2017wasserstein} between the true-unknown data distribution $P_X$ and the latent variable model $P_G$ specified by the prior distribution $P_Z$ of latent codes $Z \in \mathcal{Z}$ and the generative model $P_G(X|Z)$ of the data points $X \in \mathcal{X}$ given $Z$ \citep{tolstikhin2017wasserstein}. \eqref{eq:tolstikhin2017wasserstein} shows that while the decoder pursues the encoded training examples reconstruction at the minimal cost $c$, the encoder pursues two conflicting goals at the same time: 
\begin{enumerate*}[label=(\roman*)]
    \item Match the encoded distribution $Q_Z$ to the prior distribution $P_Z$, where $Q_Z := \mathbb{E}_{P_X} \left[Q(Z|X)\right]$
    \item Ensure that the latent representation for the decoder allows accurate reconstruction of the encoded training examples.
\end{enumerate*}
In this two steps procedure, 
first $Z$ is sampled from a fixed distribution $P_Z$ on a latent space $\mathcal{Z}$, and then $Z$ is mapped to $\hat{X} = G(Z)$ for a given map $G: \mathcal{Z} \to \mathcal{X}$, where $\hat{X} \in \mathcal{X} = \mathbb{R}^{t \times d}$.

\begin{equation}
\label{eq:tolstikhin2017wasserstein}
\begin{split}
\mathcal{L}_{\textit{WAE}}(P_X, P_G) & :=  \inf_{Q(Z|X) \in \mathscr{Q}} \mathbb{E}_{P_X} \mathbb{E}_{Q(Z|X)} \left[c(X,{G}(Z))\right] \\
& + \lambda \cdot \mathcal{D}_Z(Q_Z,P_Z), \\
& \lambda > 0
\end{split}
\end{equation}

This task formulation extends the Split-brain Autoencoder proposed by \cite{zhang2016splitbrain}. We share the intuition, and the goal of achieving a representation containing high-level abstraction and semantics of the smartphone-bus interaction registered independently by GPS and BLE sensors. In contrast with Zahng, we aim at learning the cause-effect function and its inverse, separately, and not just merely as a ``pretext''. However, to keep up with the Big Data scale, Zhang approach brings some limitations with the objective function in Eq. \eqref{eq:split-brain-Zhang}:
\begin{enumerate*}[label=(\roman*)]
    \item For weighting the multi-task cost $\mathcal{O}$, Zhang introduces the hyperparameter $\hat{\lambda}$ that requires a dedicated optimization process.
    \item To learn cause-effect relationship and its inverse, we do not want include the full signal $c((\mathcal{F}_1(X_2),\mathcal{F}_2(X_1)), X)$ in the multi-task objective function $\mathcal{O}$.  
    \item The use of a classical unregularized auto-encoder, which minimizes only the reconstruction cost $c$, between $X$ and $\hat{X}$, prevents from yielding full advantage of representation learning for this problem, facilitating model over-fitting instead of generalization power.
\end{enumerate*}

\begin{equation}
\label{eq:split-brain-Zhang}
\begin{split}
\mathcal{O}  & = arg \min_{\mathcal{F}_1, \mathcal{F}_2 \in \mathcal{F}} [ \hat{\lambda} \cdot {c}(\mathcal{F}_2(X_1), X_2) \\
& + \hat{\lambda} \cdot {c}(\mathcal{F}_1(X_2), X_1) \\
& + (1-2 \cdot \hat{\lambda}) \cdot {c}((\mathcal{F}_1(X_2),\mathcal{F}_2(X_1)), X) ], \\
& \hat{\lambda} \in [0,\frac{1}{2}]
\end{split}
\end{equation}

In the following sections we can now look at how we extended Zhang's work to cover both of the aforementioned limitations and enable clustering.

\subsection{Extension Towards Multi-task Self-learned Cost Weights}
\label{sec:multitaskExtension}
In a multi-task setting, Kendall shows that when tasks uncertainty depends on its unit of measure, homoscedastic uncertainty is an effective bias for weighting multiple losses \citep{kendall2018multi}. This fits exactly with our problem, where the proximity between smartphone and bus is measured in meters on one hand, and in Received Signal Strength Indicator (RSSI) on the other hand. With $\hat{X}_1 = \mathcal{F}_1(X_2)$ and $\hat{X}_2 = \mathcal{F}_2(X_1)$, where $\mathcal{F}_1, \mathcal{F}_2 \in \mathcal{F}$, \eqref{eq:multi-task-kendal} represents the multi-task loss formulation for our problem, according to Kendall. The main difference between \eqref{eq:split-brain-Zhang} and \eqref{eq:multi-task-kendal} is that in the second case the two parameters can be ``learned'' leveraging the ANN back propagation algorithm while learning $\mathcal{F}$ parameters, during the training phase. When training on large datasets, this is an advantage.
\begin{equation}
\label{eq:multi-task-kendal}
\begin{split}
\mathcal{O}  & = arg \min_{c} [ \frac{1}{2 \sigma_1^2} \cdot {c}(\hat{X}_1, X_1) \\
&+ \frac{1}{2 \sigma_2^2} \cdot {c}(\hat{X}_2, X_2) \\
&+ \ln{\sigma_1} + \ln{\sigma_2} ]
\end{split}
\end{equation}

\subsection{Extension towards regularized auto-encoder}
\label{sec:WAEextension}
WAE represent a class of generative models resting on the optimal transport cost derived from \cite{villani2003topics} and expressed in \eqref{eq:tolstikhin2017wasserstein}. This class underpins our extension: In contrast to Zhang work \citep{zhang2016splitbrain}, which studies the unregularized cost $c$, such as regression and cross-entropy, we include to the regression cost a regularization term, i.e., the maximum mean discrepancy (MMD) $D_Z=\textit{MMD}_k(P_Z,Q_z)$. \eqref{eq:MMD} expresses the MMD, where $k:\mathcal{Z} \times \mathcal{Z} \to \mathbb{R}$ is a positive-definite reproducing kernel, and $\mathcal{H}_k$ is the reproducing kernel Hilbert space (RKHS) of real-valued functions mapping $\mathcal{Z}$ to $\mathbb{R}$ \citep{gretton2008kernel}.

Similarly to variational auto-encoders (VAE) \citep{kingma2013auto}, this WAE-MMD formulation uses artificial neural networks (ANN) to parametrize encoder and decoder. However, to allow back-propagation throughout decoder and encoder, the re-parametrization trick \citep{kingma2013auto} \textit{``forces $Q(Z|X=x)$ to match $P_Z$ for all the different samples $x$ drawn from $P_X$. In contrast, WAE forces the continuous mixture $Q_Z : = \int Q(Z|X)dP_X$ to match $P_Z$''} \citep{tolstikhin2017wasserstein}. Consequently, WAE allow a better organization of the latent space which we leverage for clustering. Compared to alternative formulations of the penalty term, such as the Generative Adversarial Networks \citep{makhzani2015adversarial} (GAN), or in general the WAE-GAN \citep{tolstikhin2017wasserstein}, where $\mathcal{D}_{Z}$ in \eqref{eq:tolstikhin2017wasserstein} is the Jensen-Shannon Divergence, the literature shows slightly better reconstruction performance for $\hat{X}$ but at the heavy cost of an additional network and possibly complex and multi-modal distributions for $P_Z$. Since our problem is simple in principle, we opt for simplicity, thus for MMD. 

\begin{equation}
\label{eq:MMD}
\begin{split}
\textit{MMD}_k(P_Z,Q_z) &= || \int_{\mathcal{Z}} k(z,\cdot) dP_Z(z) \\
&- \int_{\mathcal{Z}} k(z,\cdot) dQ_Z(z)||_{\mathcal{H}_k},
\end{split}
\end{equation}

If $k$ is characteristic\footnote{Given $k:\mathcal{Z}^+ \to \mathbb{R}$, $k$ is injective, $\mathcal{Z}^+$ is positive and represents the set of probability measures on $\mathcal{Z^+}$} MMD represents a divergence measure \citep{sriperumbudur2011universality}. 

We try both the alternative kernels $k$ proposed for Wasserstein auto-encoders (WAE) \citep{tolstikhin2017wasserstein}: Radial basis function kernel (RBF) \eqref{eq:RBF}; and Inverse multiquadratics kernel \eqref{eq:IMK}.

\begin{equation}
\label{eq:RBF}
    k^{\textit{RBF}}(z,\Tilde{z}) =  e^\frac{-||\Tilde{z}-z||_2^2}{\hat{\sigma}_k^2}
\end{equation}

\begin{equation}
\label{eq:IMK}
    k^{\textit{IMK}}(z,\Tilde{z}) =  \frac{C}{C+||z-\Tilde{z}||_2^2}
\end{equation}

The resulting architecture consists of two independent encoder/decoder maps $\mathcal{F}_1, \mathcal{F}_2 \in \mathcal{F}$ such that $\hat{X_1} = \mathcal{F}_1(X_2)$ and $\hat{X_2} = \mathcal{F}_2(X_1)$. Each map's encoder consists of 1D-Convolutions; 1D-Transpose-Convolutions for the decoder. As described in Fig. \ref{fig:CEMWA}, maps are learned using back-propagation to minimizing the multitask formulation of our objective function \eqref{eq:CEMWA}, where we set $c = ||X-\hat{X}||_2^2$ and $D_Z=\textit{MMD}_k$. To find optimal relative weights between tasks, we leverage the same back-propagation algorithm. 

\begin{equation}
\label{eq:CEMWA}
\begin{split}
\mathcal{O}_{\textit{WAE}}  & = arg \min_{\mathcal{F}_1, \mathcal{F}_2 \in \mathcal{F}} \frac{1}{2 \sigma_1^2} \cdot \mathcal{L}_{\textit{WAE}}(\mathcal{F}_2(X_1), X_2) \\
&+ \frac{1}{2 \sigma_2^2} \cdot \mathcal{L}_{\textit{WAE}}(\mathcal{F}_1(X_2), X_1) \\
&+ \ln{\sigma_1} + \ln{\sigma_2}
\end{split}
\end{equation}

\subsection{Extension of Deep Clustering Architecture}
\label{sec:deepClustering}

To allow unsupervised classification of images, Caron et al. proposes a straight ANN predicting cluster assignment as pseudo-labels \citep{Caron_2018_ECCV}, and iterate between clustering with k-means \citep{LIKAS2003451} and back-propagation to update the network's weights after the cluster assignment. The intuition is that clustering provides and alternative and meaningful reference to labels. Therefore, the loss function is computed against clusters instead of known labels. However, since we collect two independent measure of the same event, by design, we tweak the process using these two signal as reciprocal pseudo-labels instead. When back-propagation converges, we perform clustering of data representation on the latent space with DBSCAN \citep{6814687}. Fig. \ref{fig:CEMWA}, \ref{fig:MWA} and \ref{fig:WA} show the architectures tested within our ablation study: the first leverages the known cause-effect relationship between GPS and BLE signal; the second, the multi-task independent reconstruction of the two signals; the last shares parameters within the same network, to reconstruct a tensor where multiple channels contain each available signal. 

\subsection{Final Model Formulation}
\label{sec:finalModel}
Fig. \ref{fig:CEMWA} presents the final structure of our CEMWA model, resulting from the Split-brain's architecture extensions described in Sec. \ref{sec:multitaskExtension}, \ref{sec:WAEextension} and \ref{sec:deepClustering}.

We will argue as follows: 
\begin{enumerate*}[label=(\roman*)]
    \item CEMWA has the ability of learning the cause-effect relationship between GPS and BLE signals recording smartphone-bus interactions.
    \item Learning such a relationship allows the exposure of self-validated features characterizing the BIBO status of users with respect to buses.
    \item These self-validated features allow unsupervised classification of users trajectories, where smartphones identify users and BLE devices identify buses.
    \item Alternative unsupervised architectures leveraging the correlation instead of cause/effect between the GPS and BLE signals---such as those described in Fig. \ref{fig:MWA} and \ref{fig:WA}--- are unable to to perform self-validated unsupervised BIBO classification.
    \item In case of labels noise, CEMWA significantly outperforms the most accurate supervised classifiers, such as random forest or XG-boost (extreme gradient boosting). 
    \item Regardless of the classification performance, CEMWA embodies both a data imputation and a validation mechanism, while supervised classifiers or alternative unsupervised architectures should rely on dedicated processes, such as an exponential weighted moving average for BLE or GPS imputation \citep{Osman2018}, and user validation for BIBO labels \citep{servizi2021context, servizi2020a}.
\end{enumerate*}

To substantiate our hypotheses through the following experiments, consistently, we designed and deployed a specific sensing architecture, and collected high quality ground truth.

\subsubsection{Ground truth collection, data cleansing, and preparation}

CEMWA's architecture mirrors the smartphone sensing platform we designed and deployed to track the activity of three autonomous buses operating an experimental public service in Denmark, between two extremes of the Lyngby campus where the Technical University of Denmark is located.

During operations these buses are tracked via GPS available from the bus telemetry, while test passengers recruited for the experiment are tracked via smartphones. The sensing platform collected GPS signals that both smartphones and buses generate. GPS collection was strictly limited around the operations area using a geo-fence \citep{Almomani2011}.
In the same area, we deployed $~300$ BLE devices: one on each bus and bus stop, plus one at the entrance/s of each building in the campus.

To become a test passenger, each user provided explicit agreement to terms and conditions presented in compliance with the General Data Protection Regulation\footnote{\href{https://lincproject.dk/en/become-a-test-passenger-at-dtu/}{Information provided to users before recruitement, access on 03-09-2021}}. The sensing platform supports both Android and iOS devices, and the Apps are published on GooglePlay\footnote{\href{https://play.google.com/store/apps/details?id=compute.dtu.linc}{LINC DTU at GooglePLay, access on 03-09-2021}} and App Store\footnote{\href{https://apps.apple.com/dk/app/linc-dtu/id1527389656}{LINC DTU at Appstore, access on 03-09-2021}} respectively. This project is a social science study, includes data and numbers only, is not a health science project, and does not include human biological material nor medical devices. Consequently, in Denmark, where the data collection took place, the Health Research Ethics Act provides a dispensation for notification to any research ethics committee.

When the smartphone is within the relevant geo-fence, in optimal conditions, the platform collects GPS with $\SI{1}{\second}$ resolution. Simultaneously, with the same resolution, the platform samples RSSI signal strength of BLE devices ``visible'' in the range of each smartphone.

\begin{figure}
\centering
\begin{minipage}{0.7\textwidth}
  \centering
  \includegraphics[trim={0 0.5in 0 0.7in},clip,width=\textwidth]{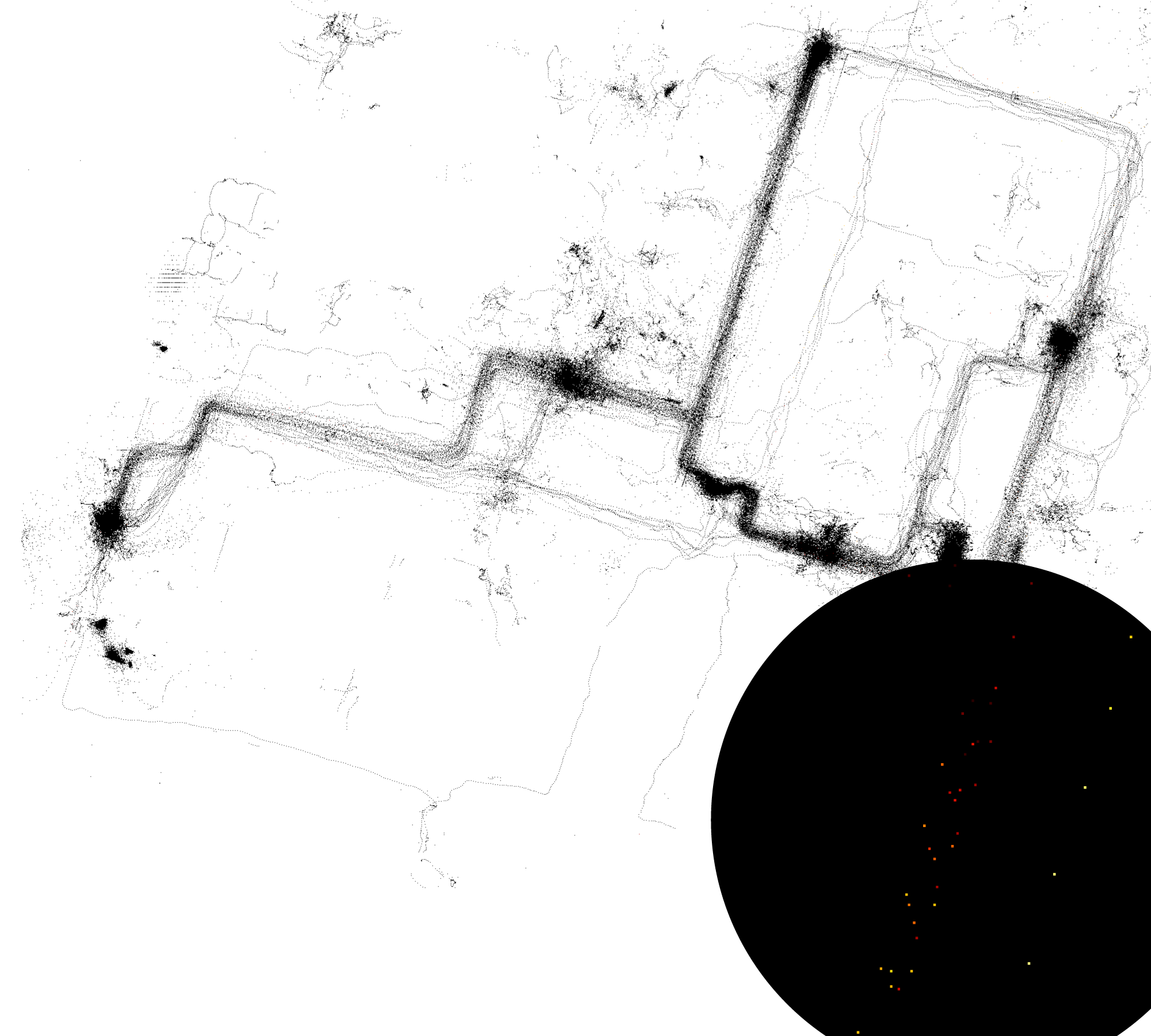}
  \vspace{-6mm}
  \caption{Subset of GPS points presenting at least one BLE device reading; color map based on $e^{speed}$ shows that buses and other modes in the area have the same speed distribution--i.e., walk and bike--few trajectories recorded from car are the only exception.}
  \label{fig:speedExp}
\end{minipage}%

\begin{minipage}{0.7\textwidth}
  \centering
  \includegraphics[trim={0 0.5in 0 0.7in},clip,width=\textwidth]{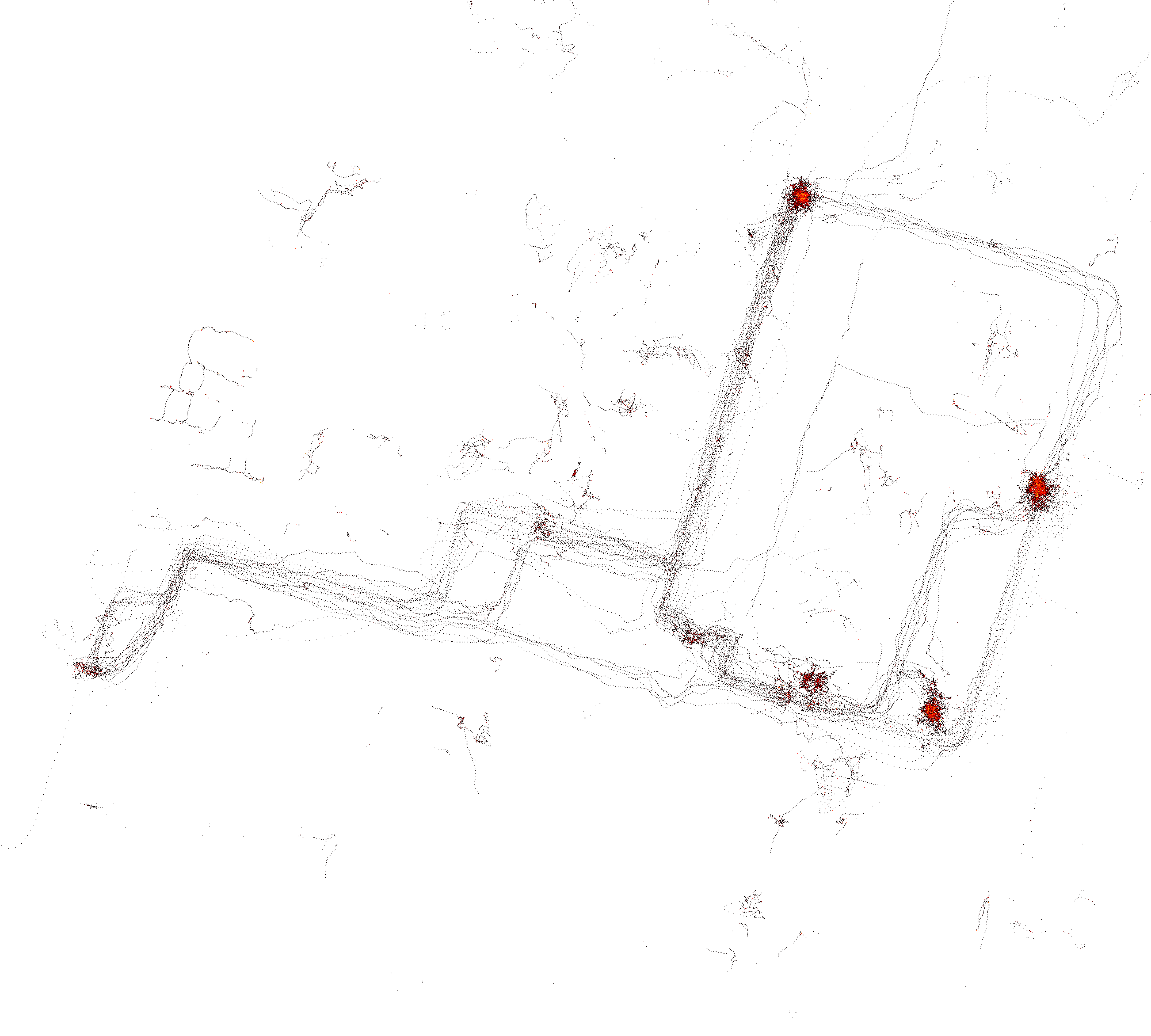}
  \vspace{-12mm}
  \caption{GPS points from smartphones, color map based on spatial density shows bus stops and bus deposit.}
  
  \label{fig:userLabelled}
\end{minipage}%
\end{figure}
\begin{figure}
\centering
\begin{minipage}{0.7\textwidth}
  \centering
  \includegraphics[trim={0 0.5in 0 0.7in},clip,width=\textwidth]{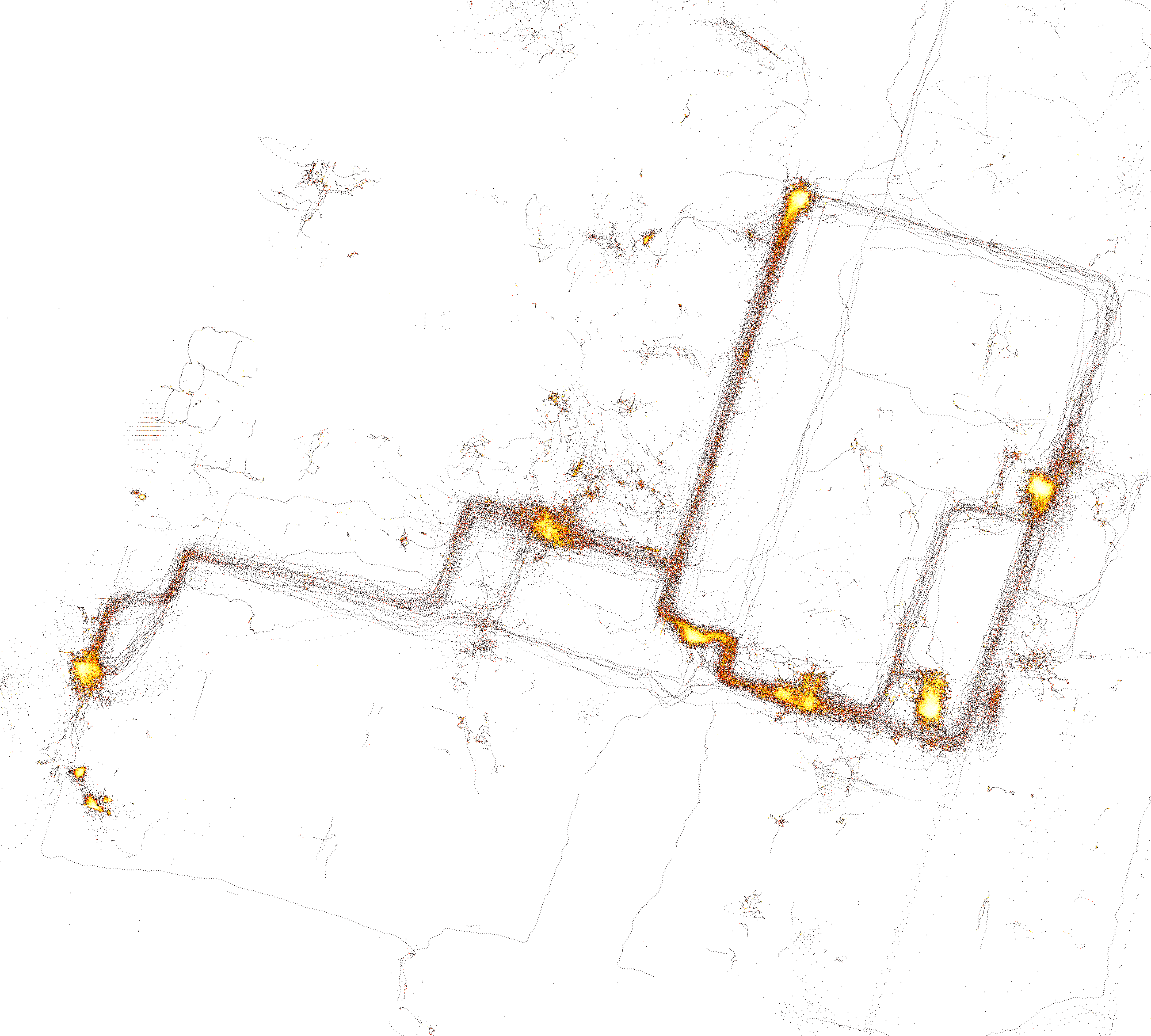}%
  \vspace{-8mm}
  \caption{Subset of GPS points presenting at least one BLE device reading; points spatial distribution shows higher density at the bus stops, bus deposit and some buildings.}
  
  \label{fig:density}
\end{minipage}

\begin{minipage}[2in]{0.7\textwidth}
  \centering
  \includegraphics[trim={0 0.5in 0 0.7in},clip,width=\textwidth]{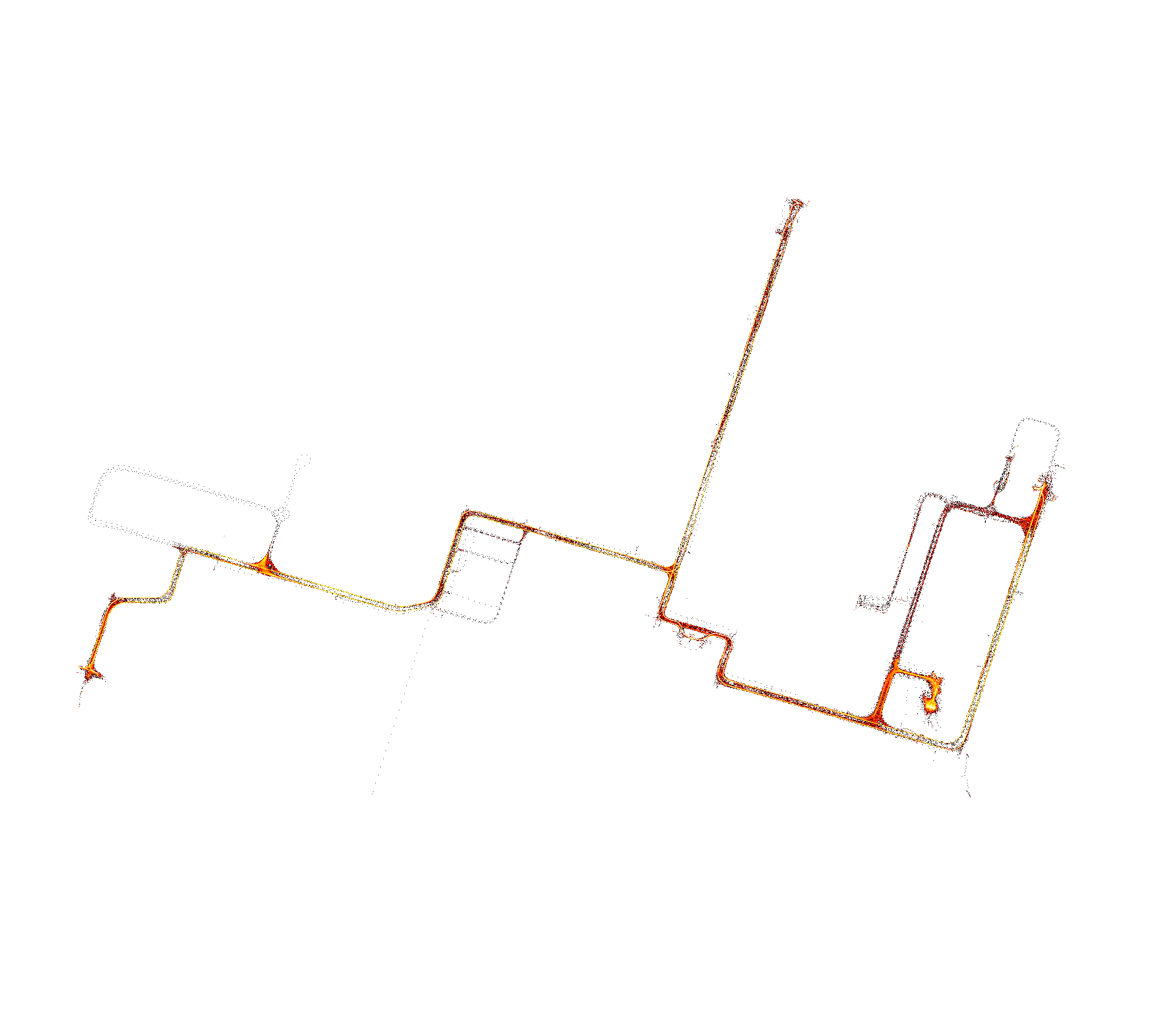}%
  \vspace{-10mm}
  \caption{GPS points from buses, spatial distribution shows higher density at the bus stops, bus deposit.}
  
  \label{fig:densityBUS}
\end{minipage}
\end{figure}
\begin{figure}
\centering
\begin{minipage}{0.7\textwidth}
  \centering
  \includegraphics[trim={0 0.5in 0 0.7in},clip,width=\textwidth]{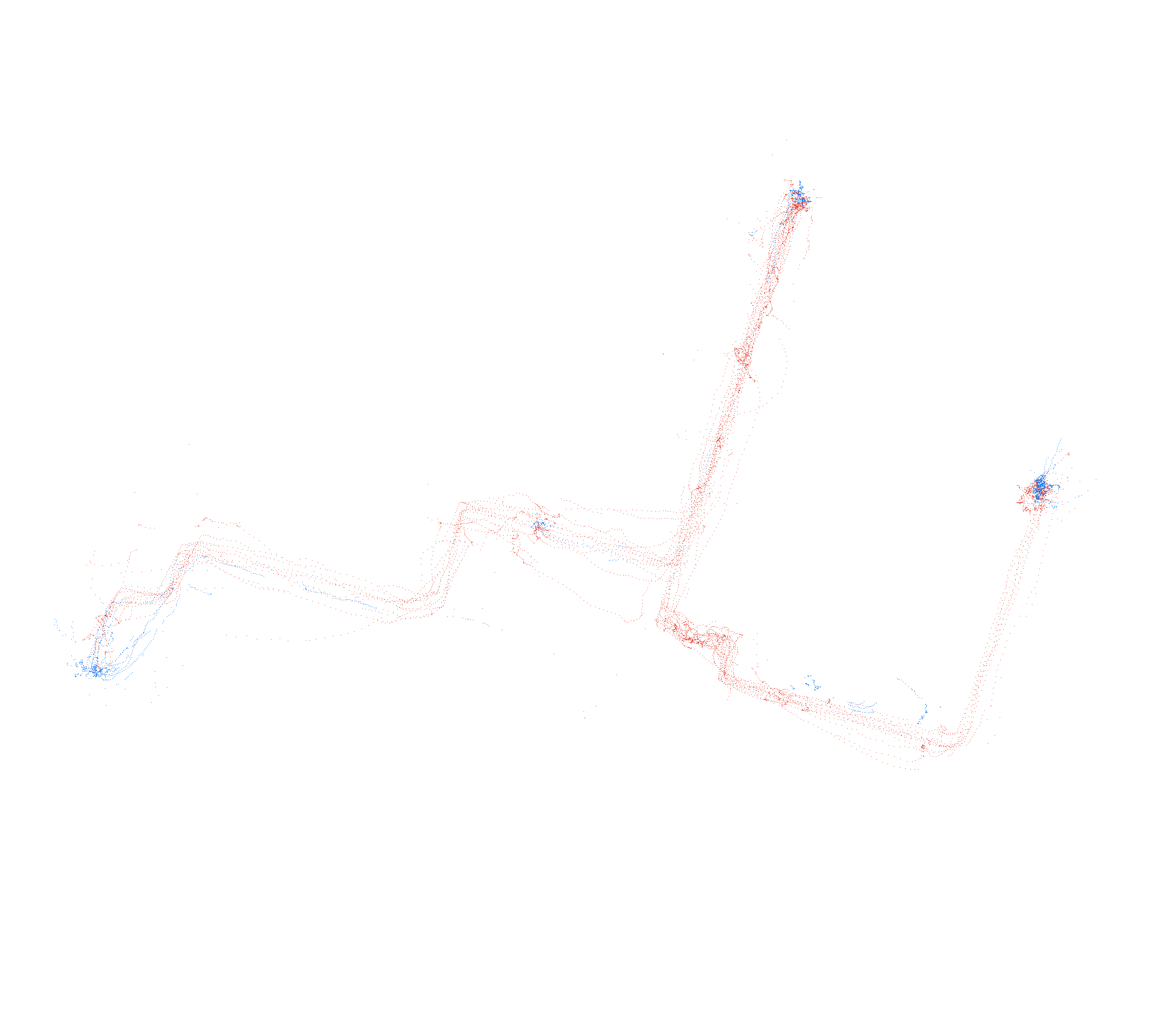}
  \vspace{-18mm}
  \caption{Be-In (BI) clusters identified on smartphone data clustering CEMWA latent space with DBSCAN, and colored with ground truth labels. Red color depicts users inside the bus; blue color, users outside the bus.}
  
  \label{fig:BI}
\end{minipage}%

\begin{minipage}{0.7\textwidth}
  \centering
  \includegraphics[trim={0 0.5in 0 0.7in},clip,width=\textwidth]{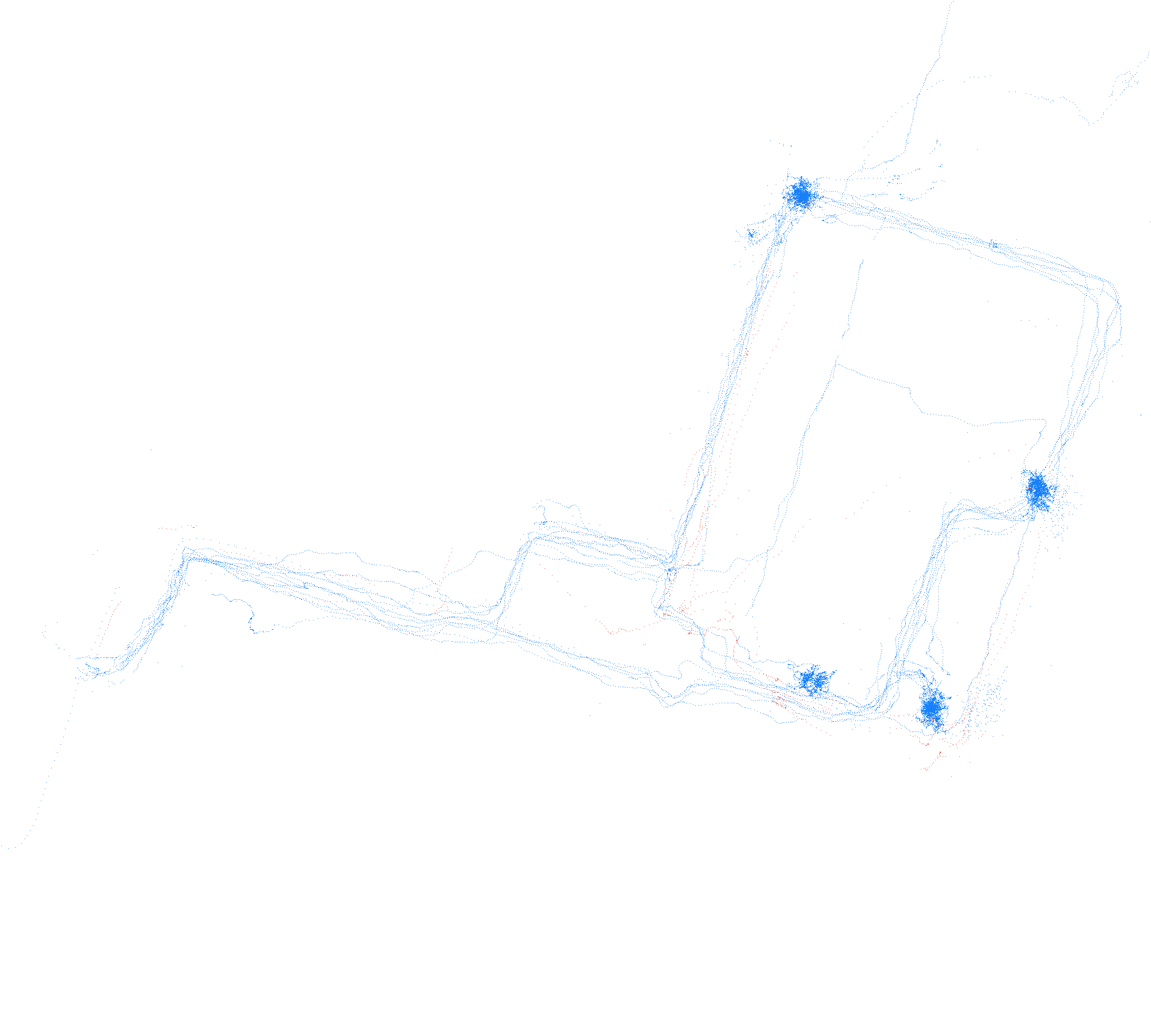}
  \vspace{-18mm}
  \caption{Be-Out (BO) clusters identified on smartphone data clustering CEMWA latent space with DBSCAN, and colored with ground truth labels. Red color depicts users inside the bus; blue color, users outside the bus.}
  
  \label{fig:BO}
\end{minipage}%
\end{figure}

We extracted the trajectories of both test passengers and buses between 1st April and 1st July. 134 users generated a total of $~4,584,000$ GPS observations; three buses, $~1,162,000$ GPS observations, for a total of approximately $\SI{940}{\hour} \cdot \text{bus}$ operations (see Fig. \ref{fig:densityBUS}).

From the remaining set of data we extracted the sub-set of observations containing at least one BLE observation, for a total of $195,000$ GPS observations (see Fig. \ref{fig:density}). This set present the maximum BLE resolution available, while the corresponding GPS resolution is below the maximum resolution available within the dataset. No labels are available for this set. Fig. \ref{fig:speedExp} depicts the speed distribution of different transportation modes present in this subset. To highlight the differences in speed between different transport mode, we applied the exponential transformation. However, the black flat color shows that the speed distribution seems to be the same in all the cases, except for some cars (see black magnified detail).

Outside the passengers' set, we generated a set of records counting $~59,000$ observations which are part of a specific experiment where seven components of the project's staff collected via smartphone a high quality BIBO labels and observations set (see Fig. \ref{fig:userLabelled}), following the same methodology of Shankari et al. for MobilityNet dataset collection \citep{shankari2020mobilitynet}. Thus, to avoid bias in the labels, we provided instructions on precise origin-destination sequences, divided in three different trip-groups. Each staff member has been randomly assigned to a trip-group. After watch synchronization, during the experiment, each staff member annotated the hour and minute each time s/he boarded or alighted a bus.


\subsubsection{Experiment setup}
Table \ref{tab:experiment-setup} describes experimental setup for the evaluation of supervised baselines, for ablation study of various unsupervised architectures, and for the model we propose in this work. We applied a trajectory segmentation considering each pair of points beyond \SI{120}{\second} time-range, or where the space variation over time variation is beyond \SI{120}{\meter/\second}, the end of a segment and the beginning of the next segment. After segmentation, for each segment we applied a sliding window including 9 consecutive points and 1 step stride. CEMWA, MWA and WA process the resulting tensor straightly, using convolutions. Instead, Random Forest and XGboost require an intermediate process to extract traditional features from the 9 step windows contained in each segment, computed at each slide, applying the same stride of 1 step. 

We setup the same conditions for both baselines and proposed methods. Comparing supervised and unsupervised classifiers in this setting is subject to the limitation of labeled dataset. As we want to provide performance distributions instead of points, with supervised methods we apply leave-one-out validation method, while with the unsupervised methods we apply a hold out method. In the first case we train the model with all the users belonging to the labeled observations except one, which represent the test set. In the test set we rotate all the users available. Thus, the main scores can presented as mean $\pm$ standard deviation. In the second case, we train the model with the unlabeled observations, and without performing DBSCAN clustering. Then we use the model including DBSCAN to classify---off the sample---the labeled observations. Similarly, we can present the main scores as mean $\pm$ standard deviation. Consequently, we can compare these scores even though the training process is quite different. 

This setup assumes that the ground truth quality is stable and high. As we mentioned, the labels collection method we used can guarantee a higher quality level on the labels. Unlike the case where ground truth is collected from passengers, the project's staff followed instructions and was not subject to, e.g., recall bias, and less likely to suffer systematic and random distractions. Therefore, to provide an exhaustive picture for performance, we train these supervised methods adding some noise in the training set, i.e., flipping a controlled percentage of labels. We sample the number of errors per user from a Poisson distribution and we flip labels accordingly. The test set is not affected. Therefore, applying a Monte Carlo evaluation based on 100 loops per experiment, and on the same setup described in Table \ref{tab:experiment-setup}, we can estimate the sensitivity to labels noise. This problem does not affect the unsupervised methods, which use Bluetooth RSSI signal as pseudo-labels instead (see Table \ref{tab:experiment-setup}, Signals row).

\begin{table*}[ht!]
\caption{Experiment Setup}
\label{tab:experiment-setup}
\resizebox{\textwidth}{!}{%
\begin{tabular}{@{}|r|c|c|c|@{}}
\toprule
\textbf{}
&
  \textbf{\begin{tabular}[c]{@{}c@{}}Supervised Baseline\\      XG-Boost\\      Random Forest\end{tabular}} &
  \textbf{\begin{tabular}[c]{@{}c@{}}Unsupervised Baseline\\      MWA (Fig. \ref{fig:MWA})\\      WA (Fig. \ref{fig:WA})\end{tabular}} &
  \textbf{CEMWA} (Fig. \ref{fig:CEMWA})\\ \midrule
\textbf{\begin{tabular}[c]{@{}r@{}}Smartphone Set\\      GPS + BLE\\      Android + iOS\end{tabular}} &
  \begin{tabular}[c]{@{}c@{}}59,000 labelled   observations\\      7 users\end{tabular} &
  \multicolumn{2}{c|}{\begin{tabular}[c]{@{}c@{}}328,000 tot   observations\\      59,000 labelled\\      134 tot users\end{tabular}} \\ \midrule
\textbf{Buses set} &
  \multicolumn{3}{c|}{\begin{tabular}[c]{@{}c@{}}1,162,000   observations,      $\SI{940}{\hour} \cdot \text{bus}$,      3 buses\end{tabular}} \\ \midrule
\textbf{Signals} &
  Speed, Longitude,   Latitude, Timestamp from GPS &
  \multicolumn{2}{c|}{\begin{tabular}[c]{@{}c@{}}Speed, Longitude,   Latitude, Timestamp from GPS\\      RSSI and Timestamp from BLE devices\end{tabular}} \\ \midrule
\textbf{Use of Ground   Truth Labels} &
  For training and evaluation &
  \multicolumn{2}{c|}{For evaluation only} \\ \midrule
\textbf{GPS Trajectory   Segmentation} &
  \multicolumn{3}{c|}{\begin{tabular}[c]{@{}c@{}}time   gap between points \textgreater \SI{120}{\second}  determines a new segment\\      
  points representing speed \textgreater \SI{45}{\meter/\second}  determine a new segment\end{tabular}} \\ \midrule
\textbf{Data Cleansing} &
  \multicolumn{3}{c|}{Segments \textless 10   consecutive points are discarded} \\ \midrule
\textbf{Observation   Imputation} &
  Imputation with   Exponential Weighted Moving Average and Masking &
  \multicolumn{2}{c|}{Masking Only} \\ \midrule
\textbf{Basic Feature   Extraction} &
  \multicolumn{3}{c|}{time-,   space-gap, and bearing between each pair of GPS points, GPS distance between   smartphone and buses within \SI{1}{\second} range} \\ \midrule
\textbf{Time Series   Sliding Window} &
  \multicolumn{3}{c|}{moving   window of 9 consecutive steps segment, and 1 step stride} \\ \midrule
\textbf{\begin{tabular}[c]{@{}r@{}}Feature Extraction \\      on Sliding Window\end{tabular}} &
  \begin{tabular}[c]{@{}c@{}}Mean value\\      Max value\\      Min value\\      Position of the minimum value\\      Position of the maximum value\\      Amplitude between min and max value\\      Number of points beyond one std dev.\\      Number of points below one std dev.\\      Number of points above one std dev.\\      Number of peaks in the mvoing window\\      Number of peaks half sliding window\\      Number of peaks above 1  one std   dev.\\      Peak distance within sliding window\\      Slope\end{tabular} &
  \multicolumn{2}{c|}{\begin{tabular}[c]{@{}c@{}} None. \\ ANN performs features extraction.\\Encoder, 1 convolutional neural network.\\Decoder, 1 transposed convolutional neural network. \\ Convolution Kernel: 3\\$\lambda \in [10^{-4},1]$ \\ Batch Size: $\in [16,1024]$\\ true sample size: $\in [10,100]$ \\ Learning Rate: $\in [10^{-5},10^{-1}]$\\ Epochs: $\in [10,100]$ \\\end{tabular}} \\ \midrule
\textbf{Performance   Evaluation Method} &
  \begin{tabular}[c]{@{}c@{}}Leave-one-out:\\      One user in the test-set\\      Training-set is the complementar set.\\      Repeated rotating each user in test-set.\end{tabular} &
  \multicolumn{2}{c|}{\begin{tabular}[c]{@{}c@{}}Hold-out: \\      Training- and validation-set from unlabelled-set.\\      Test-set corresponding to the labelled-set.\end{tabular}} \\ \midrule
\textbf{Method   performance distribution} &
  \multicolumn{3}{c|}{Given   by performance on individual users of whole the labelled set.} \\ \midrule
\textbf{Performance   Metric} &
  \multicolumn{3}{c|}{AUC ROC, F1-score, Precision, Recall, Accuracy} \\ \bottomrule
\end{tabular}
}
\end{table*}

\begin{table}[!ht]
    \caption{Encoder/Decoder CNN architecture hyperparameters, final configuration for CEMWA, EMWA, and WA.}
    \centering
    \vspace{5pt}
    \begin{tabular*}{\textwidth}{r|l}
    \hline
    \hline
         Encoder & \\ 
         Convolutional Neural Network (CNN) Layers & 1\\
         Activation Function & Rectified Linear Unit\\
         Fully connected Layers & 0\\
         Dropout & 0.25\\
    \hline
    \hline
         Decoder & \\
         Transposed CNN Layers & 1\\
         Activation Function & Leaky Rectified \\
         & Linear Unit\\
         Fully connected Layers & 0\\
         Dropout & 0.25\\
    \hline
    \hline
         Optimizer & Adam \\
         Epochs & 50 \\
         Batch Size & 32 \\
         Learning Rate & $10^-4$\\
         Dropout & 0.25\\
         
    \hline
    \hline
    \end{tabular*}
    \label{tab:HyperParCNN}
\end{table}

\section{Results and Discussion}

After a manual optimization process of CEMWA, MWA, and WA, we yield optimal performance with the combination of hyperparameters described in Table \ref{tab:HyperParCNN}. As opposed to CEMWA, MWA and WA converge to a relatively lower loss, and overfitting is higher. Although the three models have the same number of parameters, we record differing computation times for the training phase (which might be justified by concurrent processing on GPU). Compared to MWA and WA,  CEMWA achieves substantially better scores, with higher mean and inferior standard deviation. \eqref{eq:RBF} yields the results we present, while \eqref{eq:IMK} seems not effective in this use case. We apply the same penalization across all three models during back-propagation to rebalance BI and BO classes when computing the WAE loss within the optimizer. Rather than the Precision score, the Recall score of the BI class seems to provide an essential contribution to the overall superior performance of CEMWA.

The supervised methods we evaluate are performing very well. XGboost presents a slightly higher score than CEMWA but with a slightly larger standard deviation. The two models seem to have comparable performance in terms of computation time. There seems to be the following differences. In optimal conditions and ground truth quality, XGboost appears to record a substantially higher precision score, but a lower recall score than CEMWA. Under the same conditions, Random Forest seems comparable with MWA and WA, or better. But we should not forget the impact of wrong labels in the training process of supervised methods such as XGboost and Random Forest. This problem does not affect unsupervised methods like CEMWA.

\begin{figure}
\centering
\begin{minipage}{0.8\textwidth}
  \includegraphics[trim={1.5in 0.85in 1in 1in},clip,width=\textwidth]{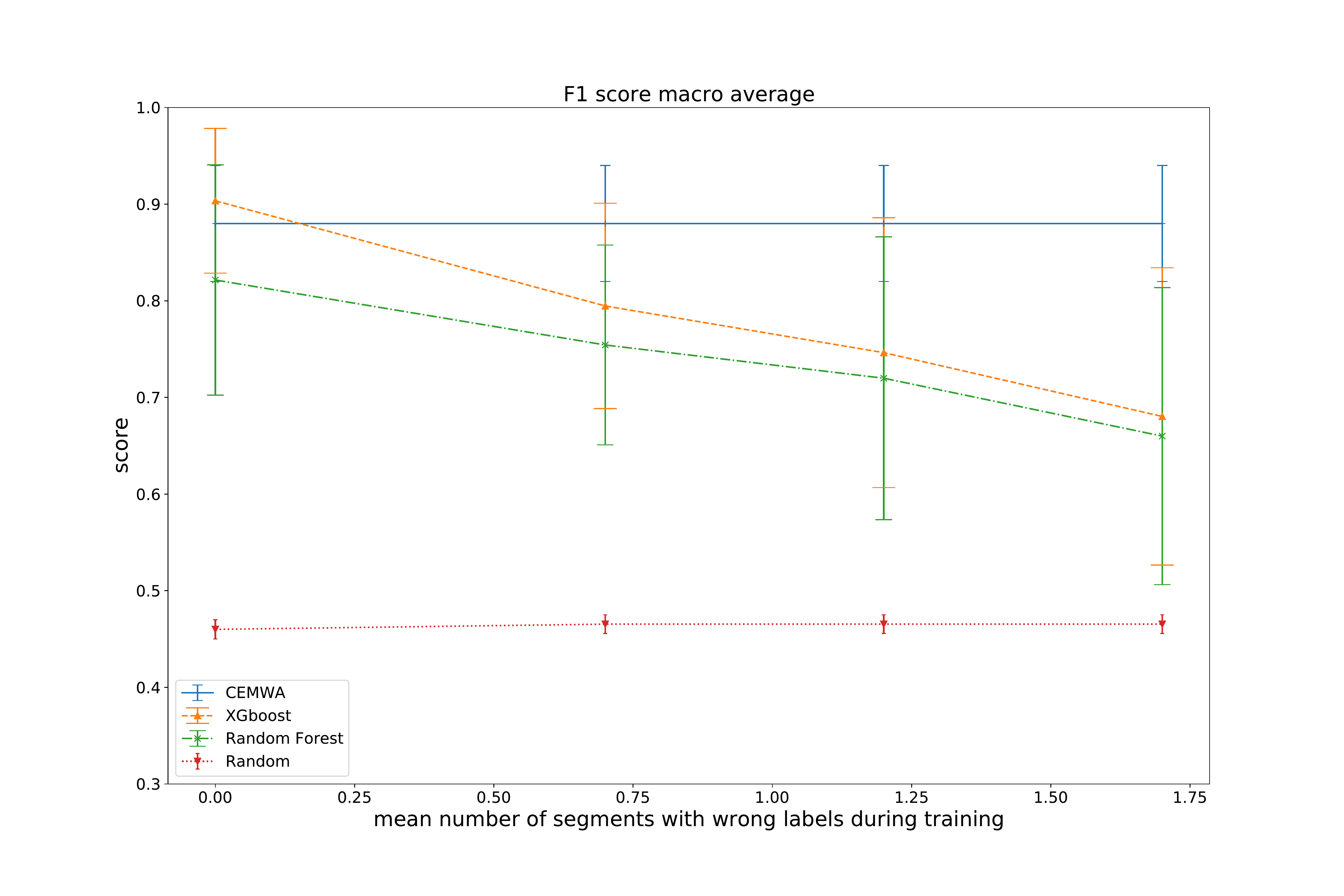}
  \caption{Impact of wrong labels on supervised classifiers training (F1 score macro average).}
  \label{fig:errorF1}
\end{minipage}%

\begin{minipage}{0.8\textwidth}
  \includegraphics[trim={1.5in 0.85in 1in 1in},clip,width=\textwidth]{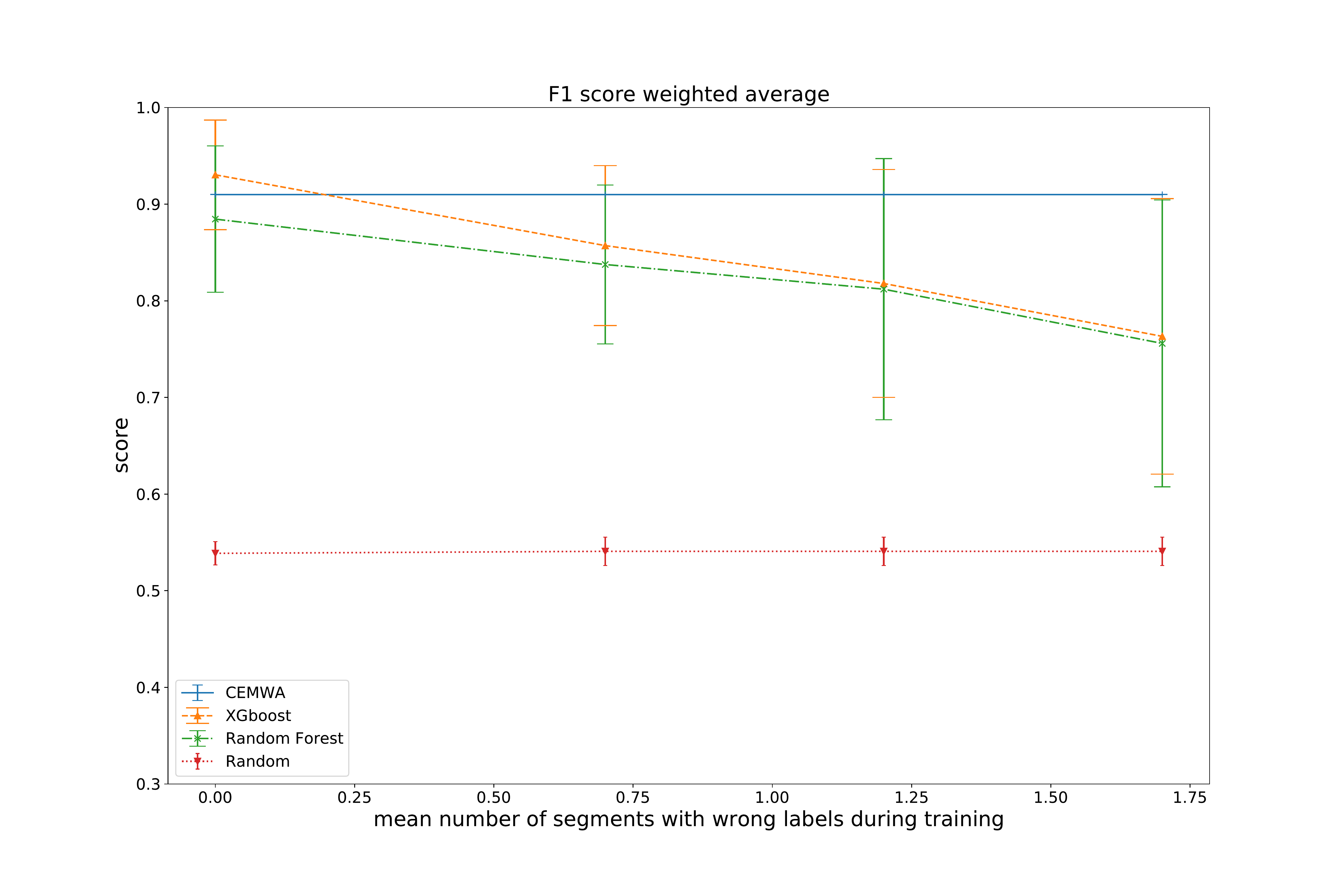}
  \caption{Impact of wrong labels on supervised classifiers training (F1 score weighted average).}
  \label{fig:errorF1w}
\end{minipage}%
\end{figure}

\begin{figure}
\centering
\begin{minipage}{0.8\textwidth}
  \centering
  \includegraphics[trim={1.5in 0.85in 1in 1in},clip,width=\textwidth]{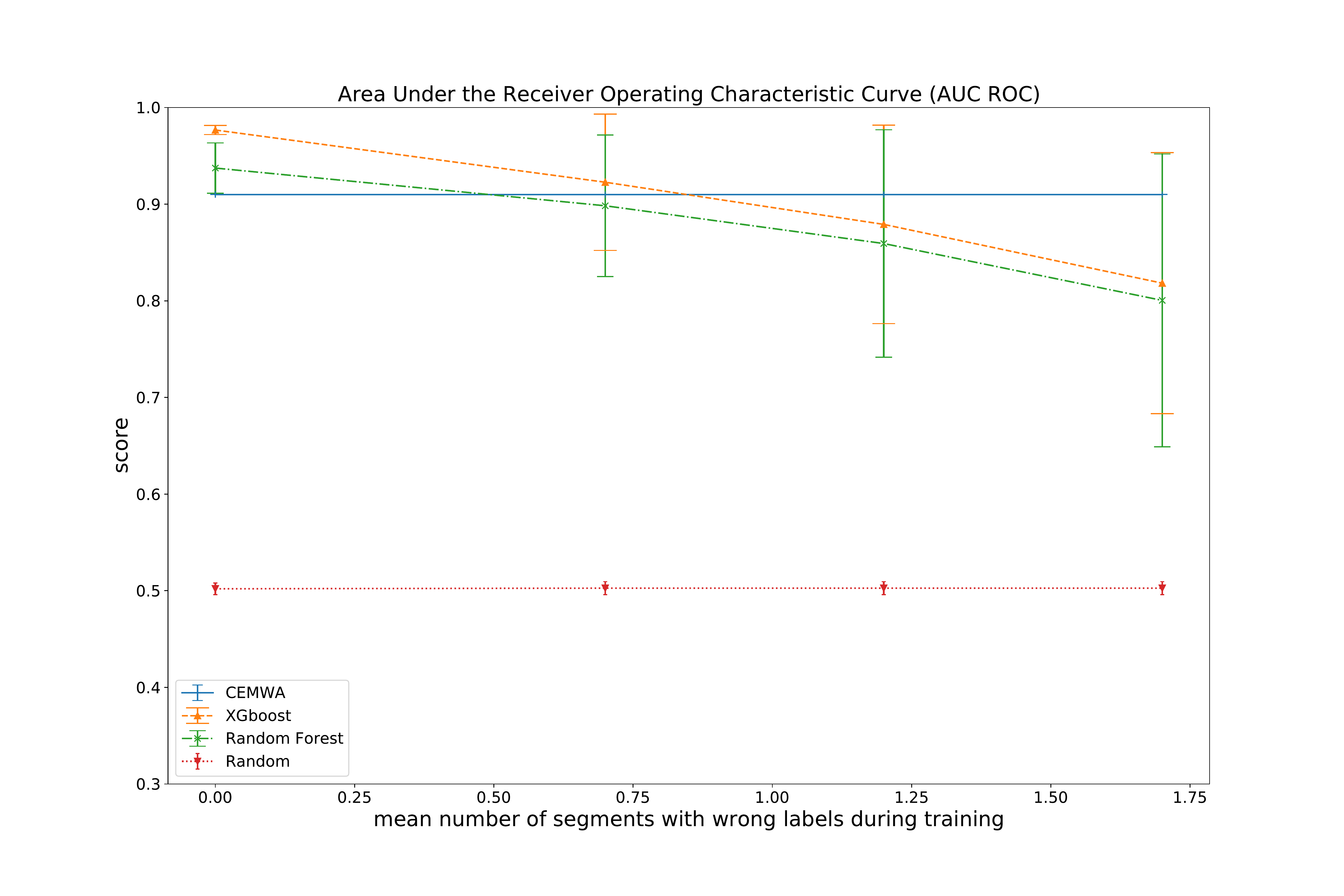}
  \caption{Impact of wrong labels on supervised classifiers training (AUC ROC).}
  \label{fig:errorAUC}
\end{minipage}%

\begin{minipage}{0.8\textwidth}
  \centering
  \includegraphics[trim={1.5in 0.85in 1in 1in},clip,width=\textwidth]{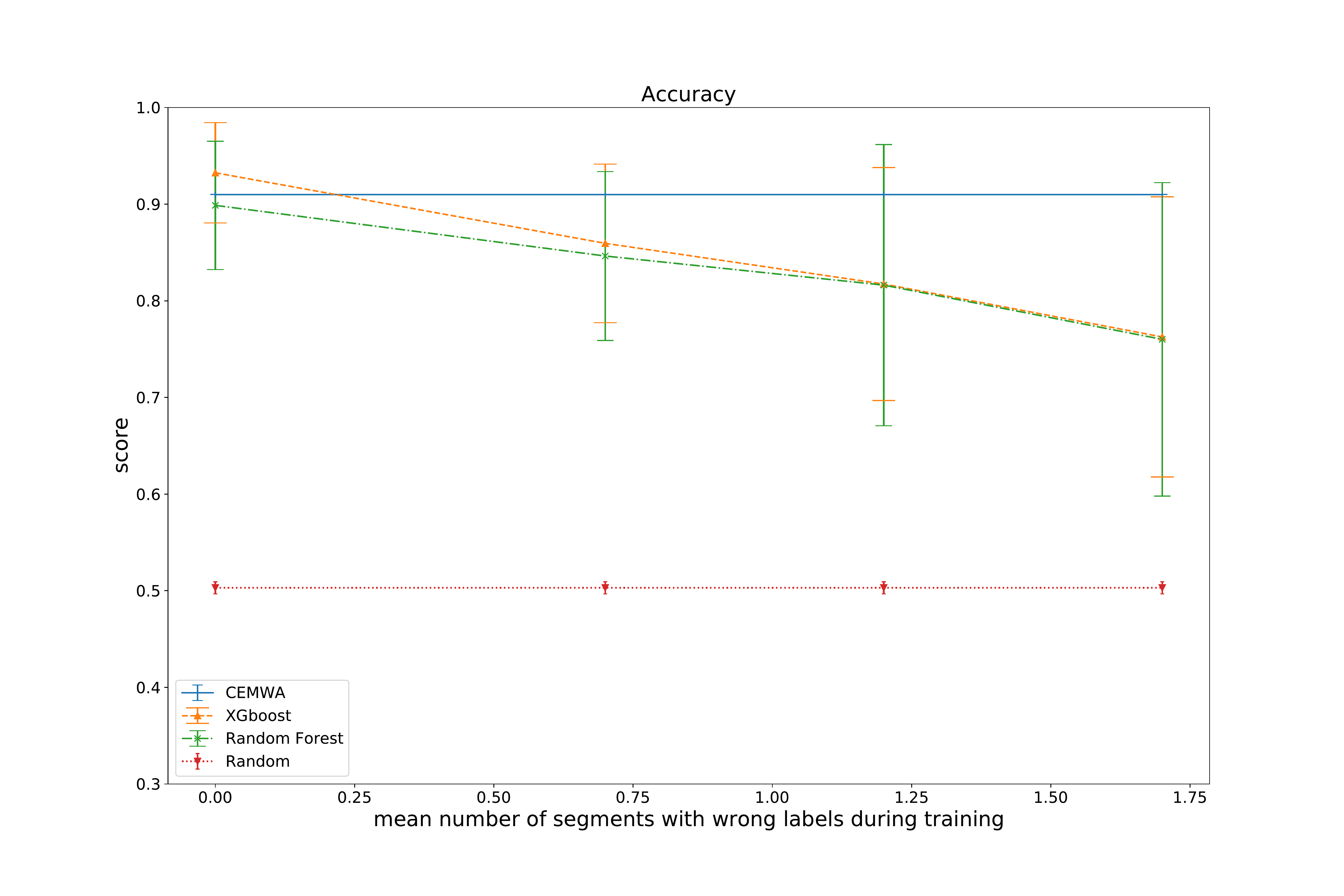}
  \caption{Impact of wrong labels on supervised classifiers training (Accuracy).}
  \label{fig:errorAccuracy}
\end{minipage}%
\end{figure}

To test the sensitivity of XGboost and Random Forest to noise in the labels, we run a Monte Carlo evaluation. Results show that beyond 10\% flipped labels during training leads to substantial performance degradation. This rapid degradation is of critical importance when labels are collected directly from passengers. Consequently, the trade-off between the cost and the quality of labels collection critically impacts the scalability potential of supervised methods. Figure \ref{fig:errorF1} depicts the impact of wrong labels on the classifiers performance: When users provide wrong labels to less than 1 segment in average--where a segment is defined according to the GPS Trajectory Segmentation of Table \ref{tab:experiment-setup}--the performance of supervised classifiers drops dramatically compared to CEMWA.

This configuration provides potential for enhancing smartphone battery efficiency and user privacy, because: 
\begin{enumerate*}[label=(\roman*)]
    \item Smartphones would listen to Bluetooth, while keeping GPS up, with minimum resolution, just enough to avoid GPS cold start;
    \item Bluetooth in proximity would trigger higher resolution GPS, only when necessary.
\end{enumerate*}

In practice, after cause-effect training with encoder-decoder architecture and clustering--where GPS compression is trained reconstructing BLE and vice-versa--CEMWA could be deployed as follows.
During operations, one CEMWA's encoder compresses GPS, while a separate encoder compress Bluetooth. The two independent compressed representation are joined into one. The proximity between the resulting representation and the clusters determine whether the observation belong to BI or BO class.

For applications where disruptions are unlikely--thus we expect a stable process in time--the amortization of high-quality ground truth could rely on a longer time horizon. An established metro line for example, is unlikely to experience changes frequently. In contrast, bus services are subject to continuous disruptions, e.g., roadworks and traffic congestion. Therefore, a supervised BIBO classifier could be a good choice in the first case. However, the unsupervised BIBO classifier seems better in the second case. Results rely mainly on the smartphone-bus-distance. This feature can be challenging to compute off-line, especially when a large number of passengers and vehicles are active. However, a federated-learning design \citep{3gpp2021} would solve the problem, and allow the computation of features online.

Assuming smartphones' future market penetration stable, and relying on adversarial sensors architectures, we show an approach to substitute manually collectible labels. This approach has vast potential; for example, BLE beacons contraposed to GPS within a CEMWA architecture would enable ticketless transit across any public transportation system, and large-scale deployment, even for applications subject to frequent disruptions. In addition to the before-mentioned use case, we suggest road and bridge tolls or sharing mobility services like cars, bikes, or scooters. A BIBO system also supports visually impaired people to chose to board the right bus from the bus stop or to alight at the right stop from the bus. It could facilitate the integration across multiple service providers, operating mostly on software instead of physical infrastructure, even integrating with existing CICO and WIWO systems. 

\begin{table*}[ht!]
\caption{Results with optimal Ground Truth for method evaluation and training of supervised algorithms}
\label{tab:results}
\resizebox{\textwidth}{!}{%
\begin{tabular}{cccccccccccccc}
\multirow{2}{*}{\textbf{Model}} &
  \multirow{2}{*}{\textbf{Task}} &
  \multirow{2}{*}{\textbf{\begin{tabular}[c]{@{}c@{}}Labeled \\ Observations\end{tabular}}} &
  \multirow{2}{*}{\textbf{\begin{tabular}[c]{@{}c@{}}Unlabeled \\ Observations\end{tabular}}} &
  \multirow{2}{*}{\textbf{Precision}} &
  \multirow{2}{*}{\textbf{Recall}} &
  \multicolumn{2}{c}{\textbf{F1-score}} &
  \multirow{2}{*}{\textbf{Accuracy}} &
  \multirow{2}{*}{\textbf{AUC ROC}} &
  \multirow{2}{*}{\textbf{\begin{tabular}[c]{@{}c@{}}Model \\ Parameters\end{tabular}}} &
  \multicolumn{2}{c}{\textbf{Computation-time}} &
  \multirow{2}{*}{\textbf{\begin{tabular}[c]{@{}c@{}}Evaluation \\ Method\end{tabular}}} \\
 &
   &
   &
   &
   &
   &
  \textbf{\begin{tabular}[c]{@{}c@{}}macro \\ average\end{tabular}} &
  \textbf{\begin{tabular}[c]{@{}c@{}}weighted \\ average\end{tabular}} &
   &
   &
   &
  \textbf{Training} &
  \textbf{\begin{tabular}[c]{@{}c@{}}Feature \\ Extraction\end{tabular}} &
   \\
\multirow{2}{*}{CEMWA} &
  BI &
  13,154 &
  \multirow{2}{*}{191,556} &
  0.77 &
  0.89 &
  \multirow{2}{*}{0.88 $\pm$ 0.06} &
  \multirow{2}{*}{0.92 $\pm$ 0.04} &
  \multirow{2}{*}{0.91 $\pm$ 0.04} &
  \multirow{2}{*}{0.91} &
  \multirow{2}{*}{5028} &
  \multirow{2}{*}{\begin{tabular}[c]{@{}c@{}}~97 min \\ on 191,556 set\end{tabular}} &
  \multirow{6}{*}{$< \SI{1}{\min}$} &
  \multirow{6}{*}{\begin{tabular}[c]{@{}c@{}}Hold out \\ (scrores \\ distribution \\ on labeled-set, \\ comparable with \\ leave-one-out)\end{tabular}} \\
 &
  BO &
  45,327 &
   &
  0.97 &
  0.92 &
   &
   &
   &
   &
   &
   &
   &
   \\

&&&&&&&&&&&&\\

\multirow{2}{*}{MWA} &
  BI &
  13,154 &
  \multirow{2}{*}{191,556} &
  0.52 &
  0.66 &
  \multirow{2}{*}{0.72 $\pm$ 0.26} &
  \multirow{2}{*}{0.79 $\pm$ 0.22} &
  \multirow{2}{*}{0.79 $\pm$ 0.26} &
  \multirow{2}{*}{0.72} &
  \multirow{2}{*}{5028} &
  \multirow{2}{*}{\begin{tabular}[c]{@{}c@{}}~47 min \\ on 191,556 set\end{tabular}} &
   &
   \\
 &
  BO &
  45,327 &
   &
  0.89 &
  0.82 &
   &
   &
   &
   &
   &
   &
   &
   \\

&&&&&&&&&&&&\\

\multirow{2}{*}{WA} &
  BI &
  13,154 &
  \multirow{2}{*}{191,556} &
  0.76 &
  0.53 &
  \multirow{2}{*}{0.77 $\pm$ 0.15} &
  \multirow{2}{*}{0.85 $\pm$ 0.08} &
  \multirow{2}{*}{0.86 $\pm$ 0.08} &
  \multirow{2}{*}{0.74} &
  \multirow{2}{*}{4932} &
  \multirow{2}{*}{\begin{tabular}[c]{@{}c@{}}~23 min \\ on 191,556 set\end{tabular}} &
   &
   \\
 &
  BO &
  45,327 &
   &
  0.87 &
  0.95 &
   &
   &
   &
   &
   &
   &
   &
   \\

&&&&&&&&&&&&\\

\multirow{2}{*}{XG-boost} &
  BI &
  13,154 &
  \multirow{6}{*}{\begin{tabular}[c]{@{}c@{}}not \\ applicable\end{tabular}} &
  0.84 &
  0.78 &
  \multirow{2}{*}{0.90 $\pm$ 0.07} &
  \multirow{2}{*}{0.93 $\pm$ 0.06} &
  \multirow{2}{*}{0.93 $\pm$ 0.05} &
  \multirow{2}{*}{0.98} &
  \multirow{6}{*}{\begin{tabular}[c]{@{}c@{}}not \\ applicable\end{tabular}} &
  \multirow{4}{*}{$< \SI{1}{\min}$} &
  \multirow{4}{*}{\begin{tabular}[c]{@{}c@{}}~31 min on \\ 58,481 set\end{tabular}} &
  \multirow{4}{*}{Leave-one-out} \\
 &
  BO &
  45,327 &
   &
  0.93 &
  0.95 &
   &
   &
   &
   &
   &
   &
   &
   \\

&&&&&&&&&&&&\\

\multirow{2}{*}{\begin{tabular}[c]{@{}c@{}}Random \\ Forest\end{tabular}} &
  BI &
  13,154 &
   &
  0.90 &
  0.43 &
  \multirow{2}{*}{0.82 $\pm$ 0.11} &
  \multirow{2}{*}{0.88 $\pm$ 0.07} &
  \multirow{2}{*}{0.90 $\pm$ 0.05} &
  \multirow{2}{*}{0.90} &
   &
   &
   &
   \\
 &
  BO &
  45,327 &
   &
  0.85 &
  0.99 &
   &
   &
   &
   &
   &
   &
   &
   \\

&&&&&&&&&&&&\\

\multirow{2}{*}{\begin{tabular}[c]{@{}c@{}}Random \\ Classifier\end{tabular}} &
  BI &
  13,154 &
   &
  0.24 &
  0.50 &
  \multirow{2}{*}{0.46 $\pm$ 0.01} &
  \multirow{2}{*}{0.54 $\pm$ 0.01} &
  \multirow{2}{*}{0.50 $\pm$ 0.003} &
  \multirow{2}{*}{0.50} &
   &
  \multirow{2}{*}{not applicable} &
  \multirow{2}{*}{not applicable} &
  \multirow{2}{*}{not applicable} \\
 &
  BO &
  45,327 &
   &
  0.76 &
  0.50 &
   &
   &
   &
   &
   &
   &
   &
  
\end{tabular}%
}
\end{table*}

\section{Conclusion}
This paper focuses on an implicit tracking system to detect whether a passenger is inside or outside the transport network. To avoid using labels in the classifier training, we leverage a novel artificial neural network architecture learning the cause-effect relationship between two independent sensors measuring the same event. We call this approach CEMWA. In optimal conditions and with high-quality ground truth, CEMWA's performance is comparable or better than both supervised and unsupervised baselines. CEMWA and XGboost performance evaluated with optimal knwoledge on BIBO ground truth seem promising for public transport ticketing in general. 
In situations with noisy ground truth--such as transport services subject to disruption or surveys where passengers lack the ticket payment as an incentive to provide exact ground truth--we show that supervised classifiers’ performance degrades. Supervised methods' tolerance to noisy labels is case specific. However, the issue does not affect CEMWA by design. Consequently, this unsupervised method is both scalable and fulfills the requirements for use-cases where, e.g., frequent service disruptions may lead to the need for regular labels' collection. Future research will investigate in few directions: 
\begin{enumerate*}[label=(\roman*)]
    \item The extension of a sensor-to-sensor validation on new signals and neural network architectures, the sensitivity to labeling noise;
    \item The introduction of sensitivity to noise as a performance index to evaluate and compare supervised methods; and
    \item The connection between dry machine learning scores of our BIBO classifier and key performance index assessing automatic fare collection systems with BIBO.
\end{enumerate*}

%

\section*{Acknowledgment}
This project is co-financed by the European Regional Development Fund through the Urban Innovative Actions Initiative.





\bibliographystyle{unsrtnat}
\bibliography{./bibtex/bib/IEEEabrv,./bibtex/bib/main}
%

%


\end{document}